\documentclass{ieeeaccess}
\usepackage{cite}
\usepackage{amsmath,amssymb,amsfonts}
\usepackage{algorithmic}
\usepackage{graphicx}
\usepackage{textcomp}

\usepackage{subfigure}
\usepackage{booktabs,siunitx}
\usepackage{multirow}
\def\BibTeX{{\rm B\kern-.05em{\sc i\kern-.025em b}\kern-.08em
    T\kern-.1667em\lower.7ex\hbox{E}\kern-.125emX}}
\begin{document}
\history{Received 12 May 2023, accepted 27 June 2023, date of publication 6 July 2023, (this is a pre-print version of the published article).}
\doi{10.1109/ACCESS.2023.3292840}

\title{Modeling the Telemarketing Process using Genetic Algorithms and Extreme Boosting: Feature Selection and Cost-Sensitive Analytical Approach}
\author{\uppercase{Nazeeh Ghatasheh}\authorrefmark{1},
\uppercase{Ismail Altaharwa \authorrefmark{2}, and Khaled Aldebei}.\authorrefmark{1}}
\address[1]{Department of Information Technology, The University of Jordan, Aqaba, Jordan (e-mail: {n.ghatasheh, k.debei}@ju.edu.jo)}
\address[2]{Department of Computer Information Systems, The University of Jordan, Aqaba, Jordan (e-mail: i\_taharwa@ju.edu.jo)}
\tfootnote{}
% \tfootnote{This paragraph of the first footnote will contain support 
% information, including sponsor and financial support acknowledgment. For 
% example, ``This work was supported in part by the U.S. Department of 
% Commerce under Grant BS123456.''}

\markboth
{Ghatasheh \headeretal: Modeling the Telemarketing Process using Genetic Algorithms and Extreme Boosting}
{Ghatasheh \headeretal: Modeling the Telemarketing Process using Genetic Algorithms and Extreme Boosting}

\corresp{Corresponding author: Nazeeh Ghatasheh (e-mail: n.ghatasheh@ju.edu.jo).}
% .

\begin{abstract}
Currently, almost all direct marketing activities take place virtually rather than in person, weakening interpersonal skills at an alarming pace. Furthermore, businesses have been striving to sense and foster the tendency of their clients to accept a marketing offer. The digital transformation and the increased virtual presence forced firms to seek novel marketing research approaches. This research aims at leveraging the power of telemarketing data in modeling the willingness of clients to make a term deposit and finding the most significant characteristics of the clients. Real-world data from a Portuguese bank and national socio-economic metrics are used to model the telemarketing decision-making process. This research makes two key contributions. First, propose a novel genetic algorithm-based classifier to select the best discriminating features and tune classifier parameters simultaneously. Second, build an explainable prediction model. The best-generated classification models were intensively validated using 50 times repeated 10-fold stratified cross-validation and the selected features have been analyzed. The models significantly outperform the related works in terms of class of interest accuracy, they attained an average of  89.07\% and 0.059 in terms of geometric mean and type I error respectively. The model is expected to maximize the potential profit margin at the least possible cost and provide more insights to support marketing decision-making.
\end{abstract}


\titlepgskip=-15pt

\maketitle

%%%%%%%%%%%%%%%%%%%%%%%%%%%%%%%%%%%%%%%%%%%%%%%%%%%%%%%%%%%
%%%%%%%%%%%%%%%%%%%% Introduction %%%%%%%%%%%%%%%%%%%%%%%%%
%%%%%%%%%%%%%%%%%%%%%%%%%%%%%%%%%%%%%%%%%%%%%%%%%%%%%%%%%%%
\section{Introduction}
\label{sec:introduction}
\PARstart{T}{elemarketing} has been the norm in social commerce activities since the revolution of social customers. Marketing teams utilize different communication mediums such as phone calls, emails, social media platforms, etc. However, it is not a switch on shift and it is challenging to predict and understand the client's behavior. Nowadays the ``social customers'' are informed and influenced by various stimuli which make the business survival more dependent on intelligent approaches. Businesses have to capture and extract the needed knowledge nuggets in understanding the market needs and make critical responses. The sought novel approaches aim far away beyond conventional market analysis practices \cite{VanLooy2022,LI2023103139}.

Knowledge is an emergent asset for building and sustaining the business ecosystem in the short and long runs. Such knowledge represents various aspects of business success factors, such as behavior, organization, culture, economic, social, and others. Benefiting from and aligning the advancements in computer intelligence to the advances in business refers to Business Analytics (BA). BA has been used by many researchers in different areas to support business creativity. For example, the behavioral analysis of service company clients to enhance customer relationship management, market segmentation for appropriate delivery of services, and enhanced customer satisfaction for a higher return on investment \cite{doi:10.1080/09537325.2021.1883583}.

BA of the banks' telemarketing process is challenging \cite{XIE2023108874,Article:Moro18,ghatasheh2020business} due to the nature of the available data and the perceptional factors of the clients. Typically, historical data about the clients are imbalanced (i.e., the ratio between clients accepting an offer and those not interested is relatively very low), missing data usually is a concern, clients' intentions and attitudes may be affected by external factors (e.g., market dynamics, economic stability, etc), and the decision-makers are interested in understanding the factors affecting the intentions of the clients \cite{moro2014data,Article:Moro18}. It is infeasible to contact all the clients in direct marketing campaigns, therefore selecting a potentially profitable sample of target clients is crucial. The cost of missing an opportunity is considerable as during the campaign either there is a possibility to exclude a potential target or contact an uninterested one. 

In practice, researchers aim at leveraging the marketing process success rate either by implementing highly demanding and complex algorithms \cite{Article:Moro18,WONG2020112918,ghatasheh2020business} (e.g., Deep Learning and Artificial Neural Networks) or employing approaches in which the original data are manipulated \cite{XIE2023108874,app11199016,YAN2020106259,li2022prediction,10.3390/math10142379} (e.g., resampling). Complex algorithms may generate competitive models but hinder the interpretation of the outcomes by decision-makers, and data manipulation may impose computational overhead or a distorted real-world representation of the problem. Either direction will make it harder to generalize the generated models and may produce analytical reports specific to the analyzed set of data; which makes predicting other clients' intentions harder to achieve.

This research adopts a novel GA-based approach to predicting the intent of bank clients for two main purposes. First, tuning the hyperparameters of a cost-sensitive classification algorithm. Second, finding the best subset of discriminating features. This approach is used to overcome various challenges of which some were imposed by the nature of the telemarketing data. The adapted cost-sensitive classification algorithm (i.e., XGBoost) managed to mitigate the effects of class imbalance without distorting the original data. Furthermore, minimizing the overall complexity of the prediction model and enabling an explainable decision-making process. The approach illustrates that simultaneous feature selection and parameter optimization is highly recommended and resulted in an outperforming prediction model, compared to various related works. The selected subset of features and optimized parameters improved significantly the performance of the prediction model such that 30\% and 10\% of the features attained 88.84\% and 89.07\% of GMean value respectively. It is worth mentioning that the approach managed to attain 0.059 as the lowest misclassification error of the positive class. 

The main contributions of this research are: 
\begin{itemize}
    \item[--] An explainable prediction model of the clients ``willing to accept a marketing offer''.
    \item[--] The significance of the dataset features.
    \item[--] Maximized potential profit margin at the least possible cost.
    \item[--] Insights to support the telemarketing decision-making process.
    \item[--] Comparison of most recent research related to bank telemarketing.
\end{itemize}

The rest of the paper is organized as follows: Section \ref{Sec_LR} is a summary of the literature related to the use of computer intelligence in telemarketing, Section \ref{dm} describes in detail the dataset used in this research and the methodology, the results and findings are discussed in Section \ref{Sec_Results}, and the main conclusions are summarized in Section \ref{Sec_conclusion}.

\section{Literature Review} \label{Sec_LR}
Vast and powerful AI advancements \cite{doi:10.1177/14707853211018428} opened promising utilization avenues in the business domain \cite{doi:10.1080/09537325.2021.1883583,Echeberria2022}. The last decades witnessed an accelerated pace of reliance on AI-based solutions in making knowledge-based decisions \cite{COOPER2023113523,Echeberria2022}. Marketing research \cite{9869838,NGAI202235} and attempts to model credit scoring \cite{tripathi2022credit} surveys are examples on the importance and potential of AI-based solutions in business domain. Telemarketing in particular is still one of the business domains that require more domain-specific solutions \cite{Echeberria2022}; the clients usually tend to be interested more in human-based interactions over computer-generated communications \cite{LI2023103139}. Automated bots or computer-generated marketing content is still behind expectations and indirect marketing may fail to convince the clients to value the marketing offers \cite{LI2023103139}. Consequently, understanding the clients' behaviors and how to predict their intentions would significantly improve the process of direct marketing (e.g., telemarketing through one-to-one phone calls). In bank telemarketing particularly, BA is an opportunity to understand the willingness of a client to accept an offer from past marketing-related data.

Moro et al. made available a rich real-world dataset containing telemarketing campaign data from a Portuguese bank~\cite{moro2014data}, which they enriched later with further national socio-economic metrics~\cite{Article:Moro18}. Since the release of the initial dataset various researchers have been attempting to generate prediction models and descriptive analytics; to assess the marketing decision-making process. However, due to the imbalance distribution of the data and other characteristics, non of the related works was able to provide solutions that handle all the challenges faced in modeling this business issue~\cite{moro2014data,9397083,Article:Moro18}. Interestingly, most of the researchers agree that not all the dataset features have the same importance in understanding the intentions of the clients~\cite{moro2014data,Article:Moro18,XIE2023108874,app11199016,YAN2020106259,doi:10.1177/0165551518770967}. Table \ref{tab_literature} describes, in brief, some of the related works and the best-attained prediction performance in terms of Geometric Mean (GMean) and Type I Error. Some of the works listed in Table \ref{tab_literature} reported the modeling performance of different approaches and Machine Learning (ML) algorithms~\cite{ghatasheh2020business,XIE2023108874,FENG2022368} and some attempted to understand the significance of the specific dataset features in predicting the willingness of the client to accept an offer~\cite{Article:Moro18,app11199016,YAN2020106259}.    

As shown in Table~\ref{tab_literature}, recent studies of bank telemarketing, approach the problem of predicting clients' willingness to enroll in term deposits from two angles, Feature selection, and imbalanced data handling. Feature selection reduces the complexity of a prediction model. Proper handling of imbalanced data allows for better generalization of the prediction model to handle real-world scenarios without exhibiting any bias towards any of the target classes. Xie et al.~\cite{XIE2023108874} proposed normalization to select features most effectively influencing a client's response to a telemarketing offer made by a bank. Further, they utilized ``SMOTE'', oversampling, to extend the minority class to overcome issues related to data imbalances. Besides losing the generality of the prediction model due to the oversampling, we argue that our approach outperforms~\cite{XIE2023108874} by digging deep to identify feature-level characteristics that help in predicting client intent. Moro, Cortez, and Rita proposed a feature selection approach based on sensitivity analysis and expert analysis~\cite{Article:Moro18}. The need for human expertise remains a raised question. Compared to the literature and our proposed solution,~\cite{Article:Moro18} make modest progress in terms of type I error. The only justification is overlooking the imbalanced nature of the data. Although features selection improved model accuracy (i.e., GMean values for the last two rows in Table~\ref{tab_literature}), this improvement was at the expense of missing clients interested in long-term deposit offers.

Safarkhani, and Moro~\cite{app11199016} proposed a combinatory approach that does feature selection and handles imbalanced data. They utilize wrapped subset evaluation to select the best discriminating features. While the proposed solution performs well in terms of GMean, it made quite poor performance in terms of type I error. In addition to this performance flaw, the way~\cite{app11199016} handles imbalanced data is questionable, as both oversampling and under-sampling are maintained simultaneously. such that correctly classified instances were increased and wrongly classified instances were reduced. Yan, Li, and Liu used ``S\_Kohonen network'' to predict the success rates of a fixed deposit by a telemarketing offer~\cite{YAN2020106259}. Their key purpose is to identify the best features to improve prediction rates of clients who are interested in long-term deposits through a telemarketing offer. By studying statistical metrics, they were able to reduce the feature space to $13$ features. Results were reported in terms of cross-check accuracy, which does not leave a window to investigate performance in terms of the different errors. Additionally, they resort to a method of bottom sampling to reduce the majority class to mitigate the issue of imbalanced data. Neglecting over $75\%$ of the publicly released data.

In~\cite{ghatasheh2020business}, Ghatasheh et al. proposed using cost-sensitive analysis to handle imbalanced data issues in the bank telemarketing domain. They achieved a pronounced improvement in terms of prediction model accuracy. They improved prediction rates of the minority class, i.e., clients intended to positively respond to the telemarketing offer, compared to the approaches of data re-sampling. In their work~\cite{ghatasheh2020business}, Ghatasheh et al. overlooked feature selection as a robust mean to reduce the complexity of prediction models. Also, they focused on the improvements to the prediction model with less discussion of the associated reflections on the economic factors. Wong, Song, and Wong investigated cost-sensitive analysis as a mean to handle imbalanced data in business applications~\cite{WONG2020112918}, They considered six business domains including churn prediction, default payment, firm fraud detection, in addition to the bank deposit telemarketing. Their proposed idea, namely CSDE, is an ensemble learning version of a cost-sensitive deep neural network. Prediction models perform well in most domains except for churn prediction, and clients' response to bank telemarketing offers. In these two domains variance between ``TPR'' and ``TNR'' exceeds $30\%$. We argue our proposed approach outperforms~\cite{WONG2020112918} in terms of modeling a client response to a telemarketing offer of a bank due to the utilization of the Genetic Algorithm (GA) as a mean for optimizing prediction model parameters.  

Gupta, Raghav, and Srivastava compared a set of boosting algorithms to the conventional machine learning techniques in the context of bank telemarketing~\cite{9397083}. They found that Light GBM outperforms all investigated classifiers in terms of accuracy, AUC score, Precision, Recall, and F1 score when considering overall data. But, by looking deep into the reported results, a key observation arises, XGBoost made very close results in most cases. It outperforms Light GBM when considering the recall rate of the positive class and the precision rate of the negative class. This observation complies with our argument that XGBoost minimizes the expected bias of the prediction model to the majority class. Besides reporting results of the investigated classifiers,~\cite{9397083} do not explicitly mention applying any means to mitigate the issue of imbalanced data or to select the best discriminating features. Feng et al. proposed a dynamic ensemble method to predict the success of a bank term deposit contract through telemarketing offer~\cite{FENG2022368}. They consider average profit in addition to the accuracy of the prediction model. They treated errors differently according to the type of misclassification. Also, they reported the important factors that impact the success of a telemarketing offer. Compared to the approach we propose here, re-weighting miss-classification resembles cost-sensitive analysis. But, here, we employ machine learning formal methods to do re-weighting. Additionally, we ignore correctly classified instances from the re-weighting process to avoid falling into over-fitting models. Besides extreme amplifying of errors,~\cite{FENG2022368}, perform under-sampling to avoid complexities brought by imbalanced data. But, they retain less than $05\%$ of the real data. In our experiments, we utilize cost-sensitive analysis to handle the issue of imbalanced data and to consider economic performance factors simultaneously. This allows us to use the whole dataset without ignoring any instance, which allows for better generalization of the prediction model.

\begin{table*}[!htb]
\centering
\caption{Summary of bank telemarketing related works.}
\label{tab_literature}
\begin{tabular}{lc p{0.2\linewidth} p{0.2\linewidth} p{0.2\linewidth} cc}
\toprule
Ref. &
  Year &
  Imbalance Handling &
  Features &
  Approach &
  GMean & Type I Err. \\
\midrule
\cite{XIE2023108874} &
  2022 &
  Oversampling (SMOTE) &
  Normalization &
  Random subspace-multi-boosting (Decision Trees C4.5) &
  94.73 & 0.073 \\
\cite{FENG2022368} &
  2022 &
  filter outliers and undersampling, Training dataset normalization &
  one-hot encoding &
  Dynamic ensemble selection (several base classifiers) &
  89.39 & 0.074 \\  
\cite{9397083} &
  2021 &
  NA &
  All features used &
  Tuned machine learning classifiers &
  78.11 &  0.337 \\
\cite{app11199016} &
  2021 &
  Resampling &
  Wrapped Subset Evaluation, Feature selection (12) &
  Decision Trees (J48) &
  87.14 & 0.211 \\
\cite{YAN2020106259} &
  2020 &
  Subsampling &
  Feature significance (13 features) &
  Optimize S\_Kohonen network &
  NA & NA \\
\cite{ghatasheh2020business} &
  2020 &
  Cost-Sensitive Classification &
  All features used &
  Cost-Sensitive Multilayer Perceptron &
  78.93 & 0.192 \\
\cite{WONG2020112918} &
  2020 &
  Cost-Sensitive Classification &
  All features used &
  Cost-Sensitive Deep Neural Network Ensemble &
  66.2 & 0.295 \\
\cite{Article:Moro18} &
  2018 &
  Subsampling and Feature Selection &
  feature relevance and expert evaluation &
  Neural Network Ensemble &
  81.79 & 0.182 \\
\cite{moro2014data} &
  2014 &
  NA &
  Relative importance, semi-automatic approach for feature selection &
  Artificial Neural Network &
  72.95 & 0.347  \\
\bottomrule
\end{tabular}
\end{table*}

%%%%%%%%%%%%%%%%%%%%%%%%%%%%%%%%%%%%%%%%%%%%%%%%%%%%%%%%%%%
%%%%%%%%%%%%%%% Data and Methodology %%%%%%%%%%%%%%%%%%%%%%
%%%%%%%%%%%%%%%%%%%%%%%%%%%%%%%%%%%%%%%%%%%%%%%%%%%%%%%%%%%
\section{Data and Methodology}
\label{dm}
This study attempts to model the process of identifying the profitable telemarketing target clients and identify the possible decision-making criteria. Starting from the approaches in the research works \cite{ghatasheh2020business} and \cite{9851666} towards a cost-sensitive feature selection approach. In telemarketing, the positive class label ``yes'' represents ``a case where a client has subscribed for a term deposit'' and the negative class label ``no'' represents the cases in which the client ``is not interested''. Various research works highlighted the advantages of cost-sensitive approaches in modeling different behaviors or decision-making processes \cite{ghatasheh2020business,ghatasheh2020cost,WONG2020112918,jiang2018cost,ling2008cost,HOPPNER2022291}. However, predicting the attitudes of the clients based on the past telemarketing data is challenging. It is inferred according to the investigated literature that an unbiased outperforming prediction model of both user classes is missing. A trade-off could be taken into consideration to favor the prediction of one of the class labels over the other without losing the overall prediction power. The cost of target misclassification depends on the incorrectly predicted class label; which is usually unequal in practice \cite{Han2019}. Therefore, directing the classifier to favor more the class of interest by increasing the weight of the positive class during the training process would reduce the potential costs. Cost-sensitive classification is expected to reduce the positive class misclassification errors while preserving an acceptable overall prediction accuracy in unbalanced datasets  \cite{ghatasheh2020business,ghatasheh2020cost}. Section \ref{dm:dataset} describes the used dataset in this research and Section \ref{dm:approach} describes the proposed approach.

\subsection{Telemarketing Dataset Description}
\label{dm:dataset}
This study considers an updated version of the real-world Portuguese bank, ``Banco de Portugal'', a benchmark marketing dataset that is available online \cite{moro2014data}. The dataset is enriched by five socio-economic national-wide indicators and it covers the period from May 2008 till November 2010.
The dataset has 41188 records of which 36548 records have a negative label and 4640 have a positive label, i.e., there are only 11\% records of the class of interest. 
The dataset consists of 20 features. Half of these features (i.e., 10 features) are numeric and the other half are categorical.
%in addition to the class label ``y''.

Table \ref{tbl_numvar} contains descriptive statistics of the numerical features.
However, Table \ref{tbl_catvar} describes the distribution of the categorical features. 
Furthermore, Table \ref{tbl_numvar} presents a descriptive statistic of the class label (i.e., ``yes'' and ``no''), which is named ``y'' in this paper.
Each numerical and categorical feature is given an index for reference.

\begin{table}[!htb]
\centering
\caption{Descriptive statistics of numerical features. (Adapted from \cite{moro2014data})}
\label{tbl_numvar}
\begin{tabular}{lcllll}
\toprule
Feature & Index &  Avg     & SD     & Min.    & Max.    \\
\midrule
age             & 0 & 40.02   & 10.42  & 17     & 98     \\
duration        & 1 & 258.29  & 259.28 & 0      & 4918   \\
campaign        & 2 & 2.57    & 2.77   & 1      & 56     \\
pdays           & 3 & 962.48  & 186.91 & 0      & 999    \\
previous        & 4 & 0.17    & 0.49   & 0      & 7      \\
emp.var.rate*   & 5 & 0.08    & 1.57   & -3.4   & 1.4    \\
cons.price.idx* & 6 & 93.58   & 0.58   & 92.201 & 94.767 \\
cons.conf.idx*  & 7 & -40.50  & 4.63   & -50.8  & -26.9  \\
euribor3m*      & 8 & 3.62    & 1.73   & 0.634  & 5.045  \\
nr.employed*    & 9 & 5167.04 & 72.25  & 4963.6 & 5228.1 \\
\bottomrule
\multicolumn{6}{l}{\textit{* Social and economic context features}}\\

\end{tabular}
\end{table}

\begin{table}[!htb]
\centering
\caption{Distribution of Categorical features and Descriptive Statistics of the class label. (Adapted from \cite{moro2014data})}
\label{tbl_catvar}
\renewcommand{\arraystretch}{0.6}% Tighter
\begin{tabular}{lclll}
\toprule
Feature                         & Index & Category             & Count & Perc. \\
\midrule
\multirow{12}{*}{job}          & 10    & admin.              & 10422 & 25.30\%  \\
                               & 11    & blue-collar         & 9254  & 22.47\%  \\
                               & 12    & entrepreneur        & 1456  & 03.54\%   \\
                               & 13    & housemaid           & 1060  & 02.57\%   \\
                               & 14    & management          & 2924  & 07.10\%   \\
                               & 15    & retired             & 1720  & 04.18\%   \\
                               & 16    & self-employed       & 1421  & 03.45\%   \\
                               & 17    & services            & 3969  & 09.64\%   \\
                               & 18    & student             & 875   & 02.12\%   \\
                               & 19    & technician          & 6743  & 16.37\%  \\
                               & 20    & unemployed          & 1014  & 02.46\%   \\
                               & 21    & unknown             & 330   & 00.80\%   \\
\midrule
\multirow{4}{*}{marital}       & 22    & divorced            & 4612  & 11.20\%  \\
                               & 23    & married             & 24928 & 60.52\%  \\
                               & 24    & single              & 11568 & 28.09\%  \\
                               & 25    & unknown             & 80    & 00.19\%   \\
\midrule
\multirow{8}{*}{education}     & 26    & basic.4y            & 4176  & 10.14\%  \\
                               & 27    & basic.6y            & 2292  & 05.56\%  \\
                               & 28    & basic.9y            & 6045  & 14.68\%  \\
                               & 29    & high.school         & 9515  & 23.10\%  \\
                               & 30    & illiterate          & 18    & 00.04\%   \\
                               & 31    & professional.course & 5243  & 12.73\%  \\
                               & 32    & university.degree   & 12168 & 29.54\%  \\
                               & 33    & unknown             & 1731  & 04.20\%   \\
\midrule
\multirow{3}{*}{\begin{tabular}[c]{@{}l@{}}default\\ \textit{credit in default.}\end{tabular}}       
                               & 34    & no                  & 32588 & 79.12\%  \\
                               & 35    & unknown             & 8597  & 20.87\%  \\
                               & 36    & yes                 & 3     & 00.01\%   \\                               
\midrule
\multirow{3}{*}{\begin{tabular}[c]{@{}l@{}}housing\\ \textit{housing loan.}\end{tabular}}
                               & 37    & no                  & 18622 & 45.21\%  \\
                               & 38    & unknown             & 990   & 02.40\%   \\
                               & 39    & yes                 & 21576 & 52.38\%  \\                               
\midrule
\multirow{3}{*}{\begin{tabular}[c]{@{}l@{}}loan\\ \textit{personal loan}\end{tabular}}
                               & 40    & no                  & 33950 & 82.43\%  \\
                               & 41    & unknown             & 990   & 02.40\%   \\
                               & 42    & yes                 & 6248  & 15.17\%  \\
\midrule
\multirow{2}{*}{\begin{tabular}[c]{@{}l@{}}contact\\ \textit{communication type}\end{tabular}}
                               & 43    & cellular            & 26144 & 63.47\%  \\
                               & 44    & telephone           & 15044 & 36.53\%  \\
\midrule
\multirow{10}{*}{month}        & 45    & apr                 & 2632  & 06.39\%   \\
                               & 46    & aug                 & 6178  & 15.00\%  \\
                               & 47    & dec                 & 182   & 00.44\%   \\
                               & 48    & jul                 & 7174  & 17.42\%  \\
                               & 49    & jun                 & 5318  & 12.91\%  \\
                               & 50    & mar                 & 546   & 01.33\%   \\
                               & 51    & may                 & 13769 & 33.43\%  \\
                               & 52    & nov                 & 4101  & 09.96\%   \\
                               & 53    & oct                 & 718   & 01.74\%   \\
                               & 54    & sep                 & 570   & 01.38\%   \\
\midrule
\multirow{5}{*}{day\_of\_week} & 55    & fri                 & 7827  & 19.00\%  \\
                               & 56    & mon                 & 8514  & 20.67\%  \\
                               & 57    & thu                 & 8623  & 20.94\%  \\
                               & 58    & tue                 & 8090  & 19.64\%  \\
                               & 59    & wed                 & 8134  & 19.75\%  \\
\midrule
\multirow{3}{*}{\begin{tabular}[c]{@{}l@{}}poutcome \\ \textit{previous marketing}\\ \textit{campaign outcome.}\end{tabular}}
                               & 60    & failure             & 4252  & 10.32\%  \\
                               & 61    & nonexistent         & 35563 & 86.34\%  \\
                               & 62    & success             & 1373  & 3.33\%   \\
% \bottomrule
\midrule[0.15em]
\multirow{2}{*}{\begin{tabular}[c]{@{}c@{}}\textit{class label.}\\\textbf{y} \end{tabular}}
& -     & no                  & 36548 & 88.73\%  \\
& -     & yes                 &  4640  & 11.27\% \\
\bottomrule
\\
\multicolumn{5}{l}{\footnotesize\textit{* Total number of records: 41188.}} \\

\end{tabular}
\end{table}

%%%%%%%%%%%%%%%%%%%%%%%%%%%%%%%%%%%%%%%%%%%%%%%%%%%%%%%%%%%
%%%%%%%%%%%%%%% The proposed approach %%%%%%%%%%%%%%%%%%%%%
%%%%%%%%%%%%%%%%%%%%%%%%%%%%%%%%%%%%%%%%%%%%%%%%%%%%%%%%%%%
\subsection{The Proposed Approach}
\label{dm:approach}
Direct marketing (e.g., phone calls and targeted emails) is one of the business opportunities that is used by many industries. However, past direct marketing data do not explicitly reveal the characteristics of potentially profitable client segments. Business-oriented organizations strive to understand the basis of the decision-making process such as in defining and contacting clients who are willing to accept an offer. Black-box solutions may offer acceptable prediction accuracy. However, they are still unable to explicitly and clearly explain the basis of a judgment. Genetic Algorithms (GA) have recently been used in simultaneous dimensionality reduction and hyperparameter optimization to tackle challenges imposed by commonly used approaches \cite{9851666}.

Marketing is still way behind the complete reliance on autonomous AI systems, human intervention is highly required at the moment \cite{LI2023103139}. Therefore, one role of AI-based solutions is to assist in understanding the clients and justifying the marketing decisions \cite{10.3390/math10142379}. There are many types of recent and advanced ML methods that are proven to attain a high level of prediction accuracy. Nonetheless, in classification problems tree-based algorithms were the most used, and gradient boosting and extreme gradient boosting (e.g., XGBoost) were the most efficient methods \cite{9869838}. In general, XGboost performs well in various domains due to its distinguished abilities in handling outliers, missing values, high dimensional data, sparse-aware processing, linear and non-linear relationships, and parallel processing. Accordingly, XGBoost is known for its scalability and ability to handle large datasets.

Marketing research, in particular, seeks an understanding of the relation between the input variables and the target class, and the level of influence on the prediction results \cite{XIE2023108874,Article:Moro18,ghatasheh2020business}. In addition, many classical ML methods seem to attract attention recently due to the vast advancement in computational resources, their ability to produce explainable models, and their simplicity \cite{9869838,10.3390/math10142379}. GA are able to compete with advanced ML methods in optimization problems where the problem space is relatively small and complex. It is apparent that most advanced methods (e.g., deep learning) are less likely to easily replace the classical ML methods in the meantime. The aforementioned characteristics would nominate a combination of a robust classification method (i.e., XGboost) and a potentially effective optimization algorithm (e.g., GA) to model the telemarketing problem.

\begin{figure}[!htb]
  \centering
  \includegraphics[width=1\linewidth]
  {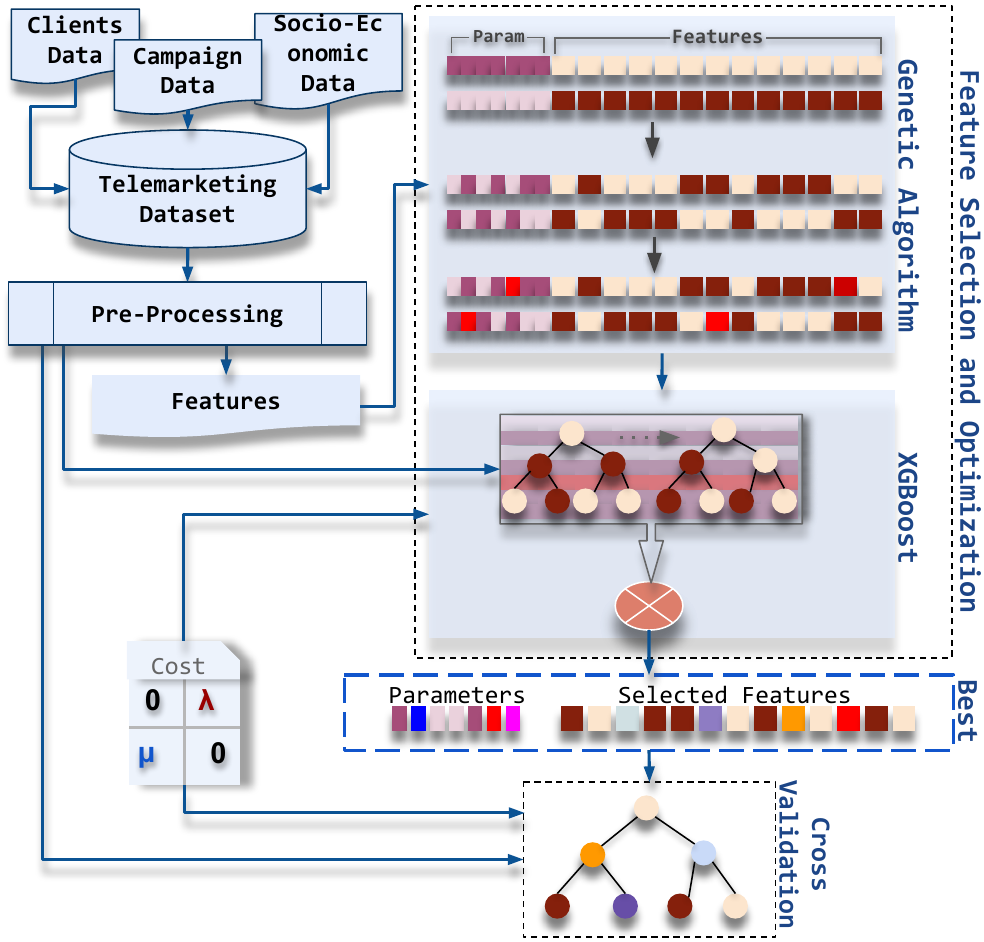}
  \caption{Abstract Architecture of the Proposed Methodology.}
  \label{fig:methodology}
\end{figure}

This research attempts to (1) develop an outperforming prediction model of a client willing to accept a term deposit, (2) find the most significant characteristics of the clients ``willing to accept an offer'', (3) maximize the potential profit margin at the least possible cost, and (4) provide more insights to support the marketing decision-making process. Figure \ref{fig:methodology} is an abstract illustration of the proposed methodology and the phases of the work are summarized as follows:
\begin{itemize}
    \item[--] Dataset preprocessing (Section \ref{subsubsec_preprocess})
    \item[--] GA feature selection and hyperparameter optimization (Section \ref{subsubsec_GAFS})
    \item[--] Analysis of the selected features (Section \ref{subsec_featureanalysis})
    \item[--] Repeated cross-validation of the best-selected models (Section \ref{subsec_crossval})
\end{itemize}

\subsubsection{Dataset Preprocessing} \label{subsubsec_preprocess}
The class label ``y'' is encoded to satisfy the requirements of some classifiers (e.g, XGBoost), i.e., the classes ``yes'' and ``no'' are replaced by ``1'' and ``0'', respectively.
The class of interest is set to be ``1'' in all of the related performance metrics. 
The marketing dataset, which is used in this research, is imbalanced and contains a mix of categorical and numerical features. Furthermore, some of the data are missed, which are denoted as ``unknown''. 
Such dataset characteristics impose many challenges that make the quality and fairness of a classification method more questionable. 
One-Hot Encoding is one of the possible solutions to alleviate these challenges \cite{doi:10.1080/1540496X.2020.1825935,Seger1259073}.
One-Hot Encoding transforms each categorical data into a new feature and assigns a binary value of ``1'' or ``0'' to those features, i.e., a categorical feature with ``m'' categories is transformed to ``m'' binary features. Each feature is assigned to ``1'' if a given data falls within the category, and zero otherwise, as illustrated in Table \ref{tbl_onehotenc}. 
In the bank marketing dataset, the 10 categorical features are transformed to 52 One-Hot Encoded features. 
Therefore, the total number of the dataset features becomes 62 features.
Table \ref{tbl_catvar} contains all the encoded features and their corresponding index in the dataset.

\begin{table}[!htb]
\centering
\caption{Three data instances under the categorical feature named ``default'' after being One-Hot-Encoded (i.e., A=no, B=yes, C=unknown).}
\label{tbl_onehotenc}
\begin{tabular}{lccc}
\toprule
Instance & default\_no & default\_yes & default\_unknown \\
\midrule
A   &   1           & 0            & 0                \\
B   &   0           & 1            & 0                \\
C   &   0           & 0            & 1               \\
\bottomrule
\end{tabular}
\end{table}

The encoded features (i.e., 52 features) make an opportunity to study the effect of each specific category on the prediction performance. 
Furthermore, feature selection may give insights into the most and least significant features. 
Detailed feature analysis and possibly an outperforming classification are facilitated by the methods described in Section \ref{subsubsec_GAFS}.

\subsubsection{GA Feature Selection and Hyperparameter Optimization} \label{subsubsec_GAFS}

In this research, further modifications are made to the GA presented in \cite{9851666} (mainly the sensitivity to the class weight and features encoding) to select a feature subset of the marketing dataset and tune the classifier hyperparameters. 
% The algorithm is modified further to make the classifier sensitive to the class weights. 
Cost-Sensitive classification \cite{WONG2020112918,9466844} aims at giving more weight to the class of interest (i.e., usually the minority class) and hence minimizing the misclassification cost. 
The weights are adjusted according to the ``Cost Matrix'' shown in Equation \ref{Eq:costMatrix}. 
The ``Cost Matrix'' represents the cost of re-weighting the classes in which $\mu$ and $\lambda$ represent the corresponding misclassification cost (i.e., False Positive and False Negative cost respectively) \cite{ghatasheh2020business,WONG2020112918,9466844}.

\begin{equation}
\mathrm{Cost}=\left[\begin{array}{cc}
0 & \lambda\\
\mu & 0
\end{array}\right]
\label{Eq:costMatrix}
\end{equation}

Confusion matrix \cite{SammutWebb2017} represents visually the classification performance.
It is considered the main base for comparison and deriving several performance metrics. 
It consists of a two-dimensional table that holds the counts of the prediction assessment (i.e., the count of correctly classified instances ``True'' and incorrectly classified instances ``False''). 
The class of interest in this study, i.e., Positive class (y=yes), represents ``Clients willing to take a term deposit'' and the Negative Class (y=no) represents ``Clients not interested in the offer''.
Therefore the counts are interpreted as follows: True Positive (TP) is the ``Clients willing to take a term deposit'' correctly classified by the algorithm as y=yes, False Positive (FP) is the ``Clients willing to take a term deposit'' incorrectly classified by the algorithm as y=no, True Negative (TN) is the ``Clients not interested in the offer'' correctly classified as y=no, and False Negative (FN) is the ``Clients not interested in the offer'' incorrectly classified as y=yes. 
The goodness of the prediction model is represented by TN and TP, and the confusion of the model (misclassification) is represented by FP and FN. The developed confusion matrix is illustrated in Table \ref{tbl:confusionMatrix}.

\begin{table}[!htb]
\centering
\caption{Confusion~Matrix (\textit{Adapted from \cite{ghatasheh2020business}}).}
\label{tbl:confusionMatrix}
\begin{tabular}{lccc}
\toprule
\multicolumn{2}{c}{} & \multicolumn{2}{c}{\begin{tabular}[c]{@{}c@{}}\textbf{Predicted}\\ take a term deposit ?\end{tabular}} \\ \midrule
\multicolumn{2}{c}{} & \multicolumn{1}{c|}{Willing to} & \multicolumn{1}{c}{Not interested} \\ \cmidrule{2-4} 
\multicolumn{1}{l}{\multirow{2}{*}{\begin{tabular}[c]{@{}r@{}}\textbf{Actual}\\ take a term deposit ?\end{tabular}}} & \multicolumn{1}{c}{Willing to} & \multicolumn{1}{c}{TP} & \multicolumn{1}{c}{FN} \\ \cmidrule{2-4} 
\multicolumn{1}{l}{} & \multicolumn{1}{c}{Not interested} & \multicolumn{1}{c}{FP} & \multicolumn{1}{c}{TN} \\ \bottomrule
\end{tabular}
\end{table}

Usually, correct classification has no incurred cost \cite{MIENYE2021100690}, i.e., C\textsubscript{TP} = C\textsubscript{TN} = 0, where $C$ represents the cost.
Machine learning algorithms consider an equal misclassification cost.
However, many business cases agree with that, i.e., classifying a fraud activity as legitimate has more drastic consequences than classifying a legitimate activity as fraud.
The same considerations apply to many other fields.
For example, the cost of diagnosing a cancer patient as healthy has more ``deadly'' consequences than diagnosing a healthy individual as a possible cancer patient \cite{MIENYE2021100690}. 
Furthermore, in telemarketing, it is expected that the cost of $C$\textsubscript{FN} is higher than the cost of $C$\textsubscript{FP} due to missing a potential opportunity to profit. 
Accordingly, Table \ref{tbl:CostconfusionMatrix} represents the desired adaptation of the misclassification cost to the confusion matrix.
Equation \ref{eq_totalcost} is used to quantify the total misclassification cost. 

\begin{table}[!htb]
\centering
\caption{Applied Cost Matrix (\textit{Adapted from \cite{MIENYE2021100690,ghatasheh2020business}}).}
\label{tbl:CostconfusionMatrix}
\begin{tabular}{lccc}
\toprule
\multicolumn{2}{c}{} & \multicolumn{2}{c}{\begin{tabular}[c]{@{}c@{}}\textbf{Predicted}\\ take a term deposit ?\end{tabular}} \\ \midrule
\multicolumn{2}{c}{} & \multicolumn{1}{c|}{Willing to} & \multicolumn{1}{c}{Not interested} \\ \cmidrule{2-4} 
\multicolumn{1}{l}{\multirow{2}{*}{\begin{tabular}[c]{@{}r@{}}\textbf{Actual}\\ take a term deposit ?\end{tabular}}} & \multicolumn{1}{c}{Willing to} & \multicolumn{1}{c}{$C$\textsubscript{TP}=0} & \multicolumn{1}{c}{$C$\textsubscript{FN}=$\lambda$} \\ \cmidrule{2-4} 
\multicolumn{1}{l}{} & \multicolumn{1}{c}{Not interested} & \multicolumn{1}{c}{$C$\textsubscript{FP}=$\mu$} & \multicolumn{1}{c}{$C$\textsubscript{TN}=0} \\ \bottomrule
\end{tabular}
\end{table}

\begin{equation}
\mathrm{Total\_Cost} = \lambda \sum \mathrm{FN} + \mu \sum \mathrm{FP}
\label{eq_totalcost}
\end{equation}

In theory, it is advised to set the cost value to the ``Inverse of the class distribution'' \cite{MIENYE2021100690}. however, in imbalanced distributions, the improper cost value may cause an extreme amplification of the positive class weight. Therefore, it is recommended to examine different cost values \cite{MIENYE2021100690,9466844} (e.g., different $\lambda$ values).
However, $\mu$ is always set to 1.
Using equation \ref{eq_inverseclassdist}, the inverse of the telemarketing dataset class distribution is computed and found to equal 7.877 ($\approx 8$). Consequently, the inverse class distribution and different arbitrary cost values were employed in this research to find a possibly suitable cost value. 

\begin{equation}
\mathrm{ClassDistributionInverse = } \frac{\sum y = \mathrm{``yes''}}{\sum y = \mathrm{``no''}}
\label{eq_inverseclassdist}
\end{equation}

The work of \cite{9851666} presented a modified GA that tries to find an optimal subset of dataset features as well as an eXtreme Gradient Boosting (XGBoost) \cite{Chen:2016:XST:2939672.2939785} hyperparameters that maximize the fitness function. 
The utilized fitness function is the Geometric Mean (GMean), which is commonly used in evaluating the prediction performance in imbalanced class distributions. 
Equation \ref{eq:gmean} shows the GMean, which is the square root of the multiplication of the positive class recall, i.e., True Positive Rate (TPR), and negative class recall, i.e., True Negative Rate (TNR) \cite{9851666,4479477,kim2015geometric,akosa2017predictive}.
The TPR and TNR are computed as shown in Equations \ref{Eq:TPR}) and \ref{Eq:TNR} respectively.

\begin{equation}\label{eq:gmean}
\mathrm{GMean}=\sqrt{\mathrm{TPR} \times \mathrm{TNR}}
\end{equation}

The GMean metric considers both classes (i.e., ``Positive'' and ``Negative'' classes) to avoid class bias in describing the model performance. 
The higher GMean indicates a better overall classification performance in predicting the two classes. 

Chen and Guestrin \cite{Chen:2016:XST:2939672.2939785} proposed a tree boosting algorithm XGBoost.
XGBoost is scalable, and it considers data compression and sharding, cache-aware access, and sparsity-aware. 
The XGBoost trains each tree on the residual error of the previous tree. 
Consequently, this results in improved performance, which is represented by the sum of the trees' predictions. 
The XGBoost is recommended by many researchers, such as \cite{9851666,MIENYE2021100690,OCCHIPINTI2022117193,Chen:2016:XST:2939672.2939785,benchm-ml2019,morde_setty_2019,nielsen2016tree}. They illustrate its outperforming behavior in many classification and regression problems if tuned properly. 

The GA initializes some parents (a.k.a. chromosomes) that consist of randomly selected classifier hyperparameters and a subset of the dataset features (i.e., chromosome genes). A set of new children (offspring) is generated through selection, uniform crossover, and mutation over various generations. GA aims at maximizing the GMean by a stochastic search for the best hyperparameters and features subset. 
The first part of the chromosomes is used to set XGBoost parameters and the rest are for the selected dataset features. 
Consequently, the best XGBoost model, which is selected based on the GMean, is employed to define the best hyperparameter values and the optimal features subset. 
The misclassification cost in XGBoost is set by the ``scale\_pos\_weight'' parameter.

\subsubsection{Repeated Cross-Validation} \label{subsec_crossval}
The GA search method is examined under different configurations (i.e., different GA setups and cost values). 
The best chromosomes are validated using 10-fold stratified cross-validation \cite{doi:10.1177/0165551518770967,9851666}.
To assert the reliability of the 10-fold cross-validation, our experiments are repeated 50 times (i.e., 50$\times$10 CV). 
The performance metrics and the selected features are reported and analyzed further to better understand the telemarketing process. 
The specific performance metrics used in the assessment are described in the next subsection (i.e., Subsection \ref{subsec_perfmetrics}).

\subsubsection{Performance Metrics} \label{subsec_perfmetrics}
The learning capability of the model development process is an essential aspect of the assessment. There are various performance metrics, derived from the confusion matrix, that describe the prediction model's ability to detect the willingness of clients to accept a marketing offer. 
Missing a potentially profitable client or contacting an uninterested client results in an incurred cost that should be minimized to gain the overall advantage.
Therefore, it is recommended to give more attention to the metrics that illustrate an overall prediction accuracy (i.e., performance metrics based on both ``Negative'' and ``Positive'' classes). 
Some of these performance metrics are briefly described as follows:

\begin{enumerate}
\item Total Accuracy (Accuracy). It is the count of correctly classified instances to the total number of instances. 
Alternatively, accuracy is referred to as success rate. 
Accuracy is calculated according to Equation \ref{Eq:acc}.

\begin{equation}
\mathrm{Accuracy} =\frac{\mathrm{TP+TN}}{\mathrm{TP+TN+FP+FN}}
\label{Eq:acc}
\end{equation}

\item True Positive Rate (TPR). It is the count of correctly classified clients willing to accept an offer (i.e., Actual ``willing to'' predicted as ``willing to'') \cite{SammutWebb2017}. 
Alternatively, TPR is named ``Recall'' or ``Sensitivity''.
TPR is calculated according to Equation \ref{Eq:TPR}.

\begin{equation}
\mathrm{TPR}=\frac{\mathrm{TP}}{\mathrm{TP+FN}}
\label{Eq:TPR}
\end{equation}

\item True Negative Rate (TNR). It is the count of correctly classified clients not interested in an offer (i.e., Actual ``not interested'' predicted as ``not interested'') \cite{SammutWebb2017}. 
Alternatively, TNR is named ``Specificity''.
TNR is calculated according to Equation \ref{Eq:TNR}.

\begin{equation}
\mathrm{TNR}=\frac{\mathrm{TN}}{\mathrm{FP+TN}}
\label{Eq:TNR}
\end{equation}

\item Type I Error. It is the False Negative Rate (FNR) as $\alpha$ (FNR = 1 - TPR) \cite{powers07evaluation}.
FNR is calculated according to Equation \ref{Eq:TypeIerr}. 

\begin{equation}
\mathrm{Type\_I\_Error} = \frac{\mathrm{FN}}{\mathrm{TP + FN}} = \mathrm{FNR}
\label{Eq:TypeIerr}
\end{equation}

\item Type II Error. It is the False Positive Rate (FPR) as $\beta$ (FPR = 1- TNR) \cite{powers07evaluation}.
Alternatively, named the probability of ``False Alarm'' or ``Fall-Out''. FPR is calculated according to Equation \ref{Eq:TypeIIerr}.

\begin{equation}
\mathrm{Type\_II\_Error} = \frac{\mathrm{FP}}{\mathrm{FP+TN}}  = \mathrm{FPR}
\label{Eq:TypeIIerr}
\end{equation}

\end{enumerate}

The Total accuracy is sometimes a misleading performance metric and particularly in imbalanced class distributions \cite{chawla2009data,ghatasheh2020cost,ghatasheh2020business,9851666}. 
Due to the high number of negative class counts compared to the positive class counts in real-life scenarios, it is recommended to support the analysis with other performance metrics. 
The classification algorithms usually have more tendency towards the negative class (i.e., bias caused by dominance) and thus less prediction accuracy of the positive class.
In imbalanced class distribution, it is usual to report the GMean and Area Under the Curve (AUC) as classification performance metrics. 
GMean and AUC are class-independent metrics and they consider the minority class in evaluating the prediction model accuracy. 
Therefore, the AUC as a metric is a representation of the diagnostic ability of the prediction model \cite{hanley1982meaning}. A smaller AUC (i.e., AUC < 0.5) indicates randomness of the classification and a higher AUC indicates better diagnosability. 
Receiver Operating Characteristic Curve (ROC) confronts the probabilities (i.e., TPR and FPR) with different threshold values. The area under ROC generates the AUC. 

In marketing research, it is common to report the Lift Curve \cite{Berry:1997:DMT:560675} of the classification model. Lift analysis shows the usefulness of the model against a random selection. Equation \ref{Eq:Lift} shows how lift is calculated at a certain sample size. 

\begin{equation}
\mathrm{Lift} = \frac{\mathrm{TPR}}{\mathrm{Sample\_Size}}
\label{Eq:Lift}
\end{equation}

Relatively high lift values mean an increased probability (lift) of selecting a positive sub-sample using the prediction model, compared to random selection.

%%%%%%%%%%%%%%%%%%%%%%%%%%%%%%%%%%%%%%%%%%%%%%%%%%%%%%%%%%%
%%%%%%%%%%%%%%% Results and Discussion %%%%%%%%%%%%%%%%%%%%
%%%%%%%%%%%%%%%%%%%%%%%%%%%%%%%%%%%%%%%%%%%%%%%%%%%%%%%%%%%
\section{Results and Discussion} \label{Sec_Results}
\begin{figure}[!htb]
  \centering
  \includegraphics[width=1\linewidth]{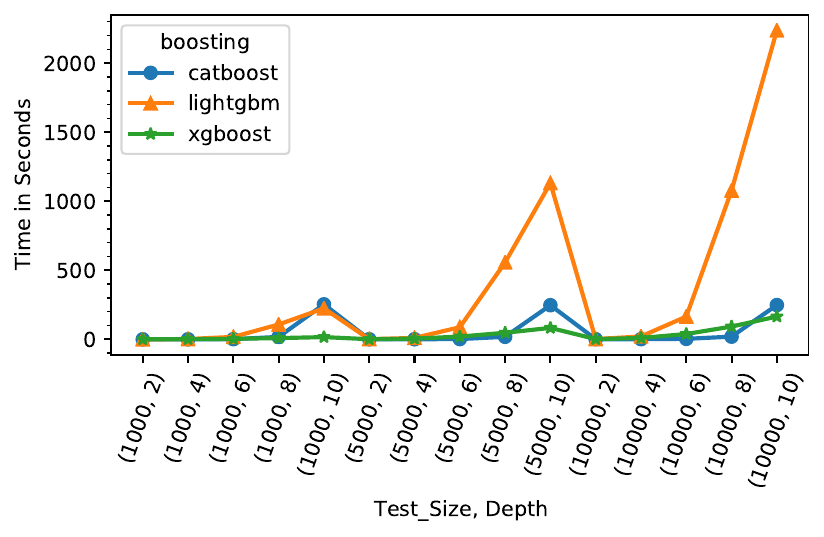}
  \caption{Testing time comparison of XGBoost, CatBoost, and LightGBM.}
  \label{fig:perfComp}
\end{figure}

The behavior of the recently used boosting algorithms XGBoost, CatBoost, and LightGBM depends on the experiment configurations. The boosting algorithms in terms of testing time are examined under different configurations. Mainly the testing time of models trained over 1000, 5000, and 10000 instances. To ensure the fairness of the comparison the same samples were used in building all the models from the Portuguese marketing dataset. For each input size, the depth of the model tree is set to 2, 4, 6, 8, and 10. Figure \ref{fig:perfComp} illustrates the testing time in seconds of the three boosting algorithms running over a CPU-based environment. It is apparent that LightGBM testing time increases significantly in relation to the depth of the model and the sample size. At the first glance, the LightGBM appears to suit the small size of telemarketing datasets and shallow tree depth.

\begin{figure}[!htb]
  \centering
  \includegraphics[width=1\linewidth]{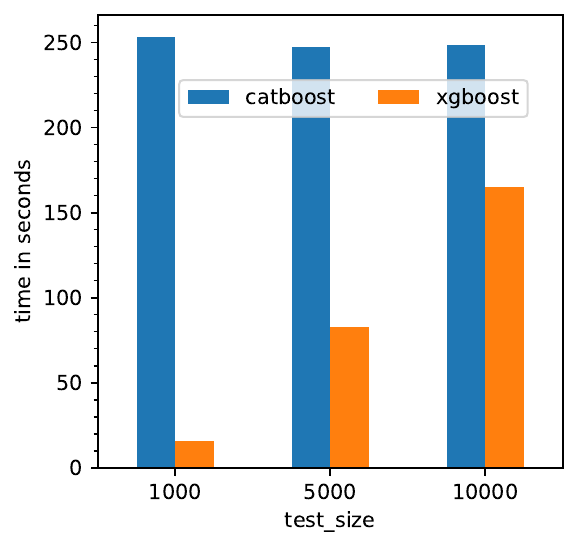}
  \caption{Testing time comparison of XGBoost and CatBoost at model depth = 10.}
  \label{fig:perfComp2depth10}
\end{figure}

As the depth of the model increases as the performance varies significantly among the three boosting algorithms. Figure \ref{fig:perfComp2depth10} illustrates the sample test size in relation to the testing time at the maximum examined depth (i,e., 10). The analysis of the testing time shows that XGBoost outperforms the confronted algorithms in larger datasets or deeper models. 

\subsection{Sensitivity Analysis}
Sensitivity analysis \cite{SensAnSaltelli2002} is an important step in building a machine learning-based model. It aims at studying the behavior of the used methods when applied to a specific case and defining the proper setup parameters. In this study, it is important to analyze the effect of the GA configurations \cite{9851666} and different cost values ($\lambda$) on the fitness function. Therefore, several GA parameters and $\lambda$ values are examined as listed in Table \ref{tbl:gafilenamecodes}. Wherever there is a random number generation method the ``seed'' is fixed (i.e., randomness seed = 723) to ensure consistent behavior in the comparisons and reproducible experiments.

\begin{table}[!htb]
\caption{GA Parameters configurations.}
\label{tbl:gafilenamecodes}
\centering
\begin{tabular}{ll}   %{p{3.4cm}p{2.9cm}}  %{ll} %{\tblwidth}{@{} LL@{} }
\toprule
Parameter  & Value                              \\
\midrule
Percent of features to select (F)  & [10,20,30,40] \%     \\
Parents (P) & [10, 20, 100]\\
Crossover ratio (C) &   [10, 20, \dots ,90] \% \\
Generations (G) &  [100, 50, 30]             \\
Cost ($\lambda$) & [1,2, \dots, 12, 15, 20, 50, 100, 150, 200] \\
\bottomrule
\end{tabular}
\end{table}

Specifically for XGBoost, the recommended ranges of the hyperparameters to be optimized are listed in Table \ref{tbl:xgboostparams} with a brief description of each parameter. The GA algorithm considers these specified ranges in optimizing the hyperparameters of XGBoost.

\begin{table}[!htb]
\caption{The hyperparameters of XGBoost to be optimized by GA \cite{9851666}.}
\label{tbl:xgboostparams}
\centering
\begin{tabular}{p{2cm}p{1.2cm}p{4cm}}
\toprule
XGB Parameter  & Value Range     & Description \\
\midrule
learning\_rate &  0.01 - 1  & Algorithm learning rate; the lower the better but requires more iterations to find an optimal solution.     \\
n\_estimators      & 10 - 1500 &  Maximum number of estimators         \\
max\_depth   &   1 - 10    & Maximum tree depth, to control overfitting. (e.g., high depth will bias the algorithm toward a specific sample)         \\
min\_child\_weight &  0.01 - 10.0  &  Minimum sum of observations in a child         \\
gamma     &    0.01 - 10      & Minimum reduction of loss when splitting        \\
subsample     &  0.01 - 1.0    & Random subset of observations for each tree           \\
colsample\_bytree & 0.01 - 1.0 & Subset of columns to be sampled in the trees         \\
\bottomrule
\end{tabular}
\end{table}

\begin{figure*}[!htb]
  \centering
  \subfigure[$\lambda \in$ \text{[1,\dots,5]}]{\label{fig:costA}\includegraphics[width=0.45\textwidth]{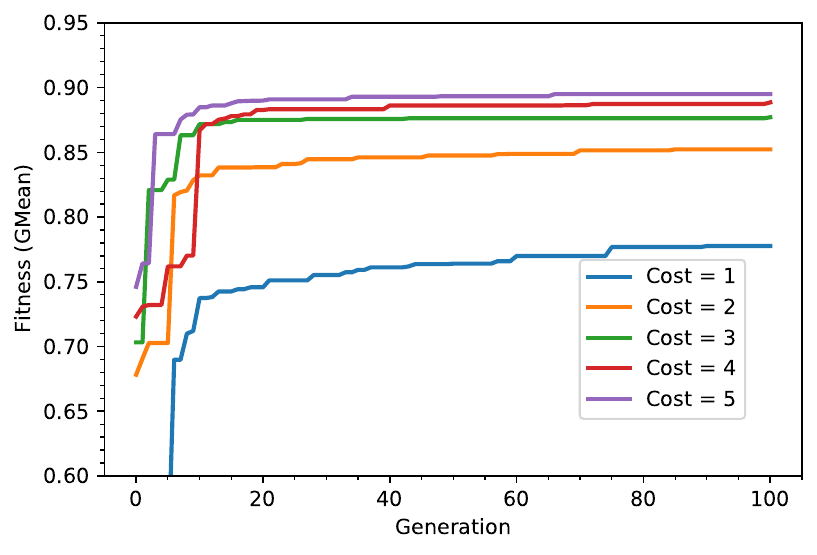}}
  \hspace{1cm}
  \subfigure[$\lambda \in$ \text{[6,\dots,10]}]{\label{fig:costB}\includegraphics[width=0.45\textwidth]{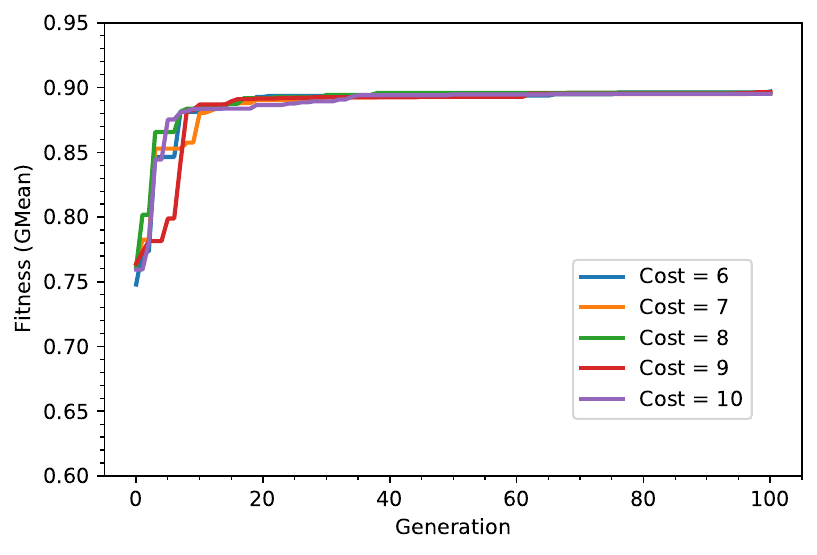}}
  \hspace{1cm}
  \subfigure[$\lambda \in$ \text{[11, 12, 15, 20]}]{\label{fig:costC}\includegraphics[width=0.45\textwidth]{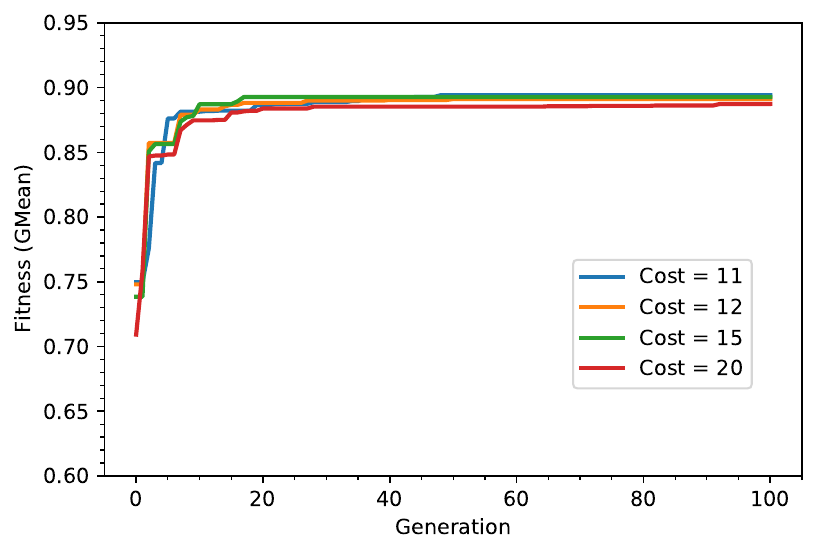}}
  \hspace{1cm}
  \subfigure[$\lambda \in$ \text{[50, 100, 150, 200]}]{\label{fig:costD}\includegraphics[width=0.45\textwidth]{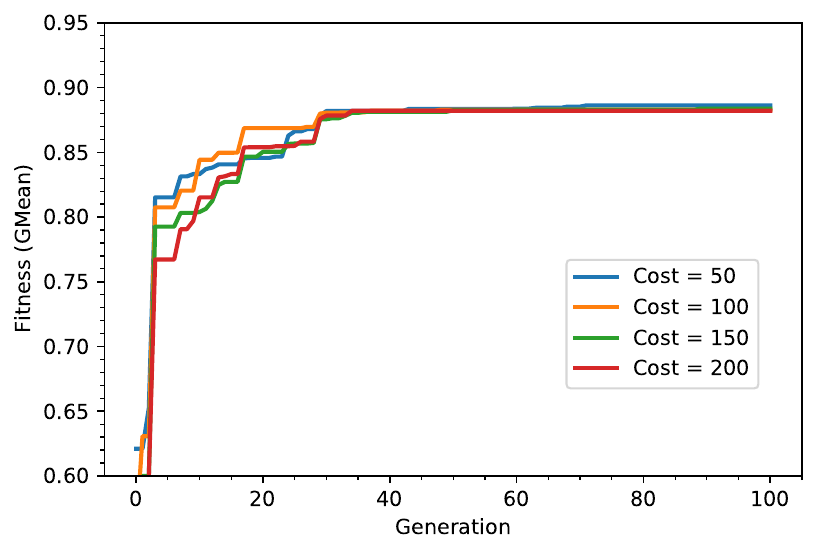}}
  \caption{The effect of different cost values on the GMean; F=10, P=10, C=5\%, G=100.}
  \label{fig:figCost}
\end{figure*}

Different cost value ``$\lambda$'' is used to re-weight the misclassification cost in different experiments. The other GA configuration parameters are fixed; the percentage of features to be selected (F) is 10, the initial population size (P) is 10, the crossover ratio (C) is 5\%, and iterated over 100 generations (G). The effect of the different cost values on the GMean is illustrated in Figure \ref{fig:costA} ($\lambda = 1, \dots, 5$ ), Figure \ref{fig:costB} ($\lambda = 6, \dots, 10$ ), Figure \ref{fig:costC} ($\lambda = 11,12, 15, 20$), and Figure \ref{fig:costD} ($\lambda = 50, 100, 150, 200$). 

\begin{figure*}[!htb]
  \centering
  \subfigure[Crossover $\in$ \text{[10\%,20\%\dots,50\%]}]{\label{fig:crossEfitE}\includegraphics[width=0.45\textwidth]{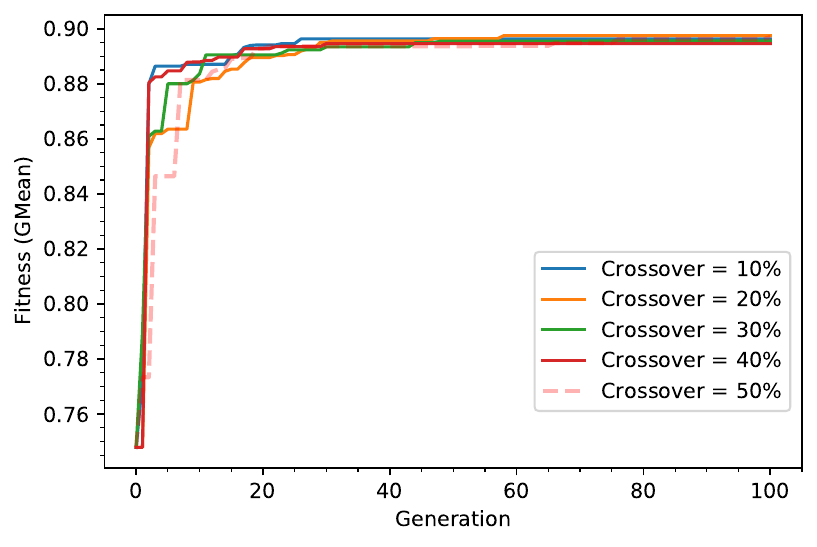}}
  \hspace{1cm}
  \subfigure[Crossover $\in$ \text{[50\%,60\%\dots,90\%]}]{\label{fig:crossEfitF}\includegraphics[width=0.45\textwidth]{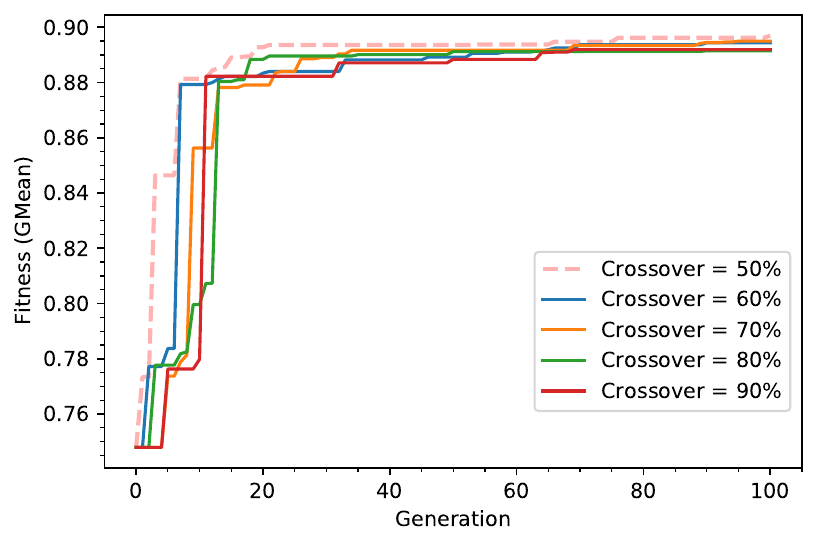}}
  \caption{The effect of different crossover ratios on the GMean; F=10, P=10, G=100, and $\lambda$=6.}
  \label{fig:figCrossOver}
\end{figure*}

It is noticeable that cost values around the inverse of the class distribution ($7.877 \thickapprox 8$), Figure \ref{fig:costB},  results in the best GMean values. It leads to better convergence of the fitness function too. The value ``$\lambda = 6$'' is the most suitable cost value for the bank telemarketing case, it is less likely to cause an amplification of the positive class weight, and it is expected to avoid overfitting in the generated models. Therefore, a cost value = 6 is fixed in the following experiments to study the effect of the crossover ratio in modeling the telemarketing process. The other GA parameters are fixed to focus on the effect of the crossover. The effect of the crossover ratio between $10\%$ and $50\%$ is illustrated in Figure \ref{fig:crossEfitE}, and Figure \ref{fig:crossEfitF} shows the effect of the crossover ratio between $50\%$ and $90\%$.

50\% Crossover ratio represents a borderline between two sets of results, (i.e., High and Low fitness). A crossover ratio below 50\% results in a higher GMean compared to a crossover over 50\%. The nature of the GA and the relatively small number of features to be selected require a small crossover proportion. In addition, the randomness of the GA in initializing the population seems to be sufficient and hence requires a low rate of crossover. The analysis of the crossover effect as illustrated in Figure \ref{fig:figCrossOver} and the cost value effect shown in Figure \ref{fig:figCost} makes it apparent that fast convergence occurs usually before 20 generations, and in general, there is no significant increase in GMean value after 30 generations.

The initial sensitivity analysis results suggest that the near-optimal configuration values are 6 as cost value, 20\% crossover rate, and 20 generations. Based on these findings a number of consequent optimization and feature selection experiments were conducted, of which some were repeated roughly with configurations according to Table \ref{tbl:gafilenamecodes}. The configurations and the results of the top ten experiments (based on GMean) are listed in Table \ref{tbl_A-JGA}. The experiments are named by letters from ``A'' to ``J'' and ``A-J'' refers to the best 10 GA feature selection and hyperparameter optimizations experiments.

\begin{table*}[!htb]
\centering
\caption{Performance summary of the top 10 GA optimization and feature selection experiments (Experiments with IDs ``A-J'').}
\label{tbl_A-JGA}
\begin{tabular}{l p{4em} p{3em} p{3em} p{3em} p{3em} p{3em} p{3em} p{3em} p{3em} p{3em} }
\toprule
Experiment:              & A        & B        & C        & D        & E       & F       & G        & H        & I        & J       \\
%\cmidrule(lr){1-1}
\cmidrule(r){2-2}
\cmidrule(r){3-3}
\cmidrule(r){4-4}
\cmidrule(r){5-5}
\cmidrule(r){6-6}
\cmidrule(r){7-7}
\cmidrule(r){8-8}
\cmidrule(r){9-9}
\cmidrule(r){10-10}
\cmidrule(r){11-11}

& \multicolumn{8}{c}{GA Setup} \\
\cmidrule{2-11}
Perc. to Sel.           & 30\%     & 30\%     & 10\%     & 30\%     & 10\%    & 10\%    & 20\%     & 40\%     & 10\%     & 10\%    \\
Parents            & 100      & 100      & 20       & 100      & 10      & 10      & 50       & 20       & 10       & 10      \\
Crossover          & 20\%     & 80\%     & 70\%     & 70\%     & 20\%    & 50\%    & 80\%     & 20\%     & 50\%     & 50\%    \\
Generations        & 30       & 50       & 100      & 30       & 100     & 100     & 30       & 30       & 100      & 100     \\
$\lambda$ & 6        & 8        & 6        & 6        & 6       & 6       & 8        & 6        & 8        & 10      \\
\\
& \multicolumn{8}{c}{Performance} \\
\cmidrule{2-11} \\
GA Time in sec.         & 6163.667 & 9807.685 & 1171.609 & 3713.541 & 937.327 & 722.467 & 2306.898 & 1021.876 & 1028.308 & 711.473 \\
Fitness (GMean)        & 90.21\%  & 89.97\%  & 89.83\%  & 89.81\%  & 89.75\% & 89.70\% & 89.68\%  & 89.65\%  & 89.60\%  & 89.52\% \\

\\
 & \multicolumn{8}{c}{Optimized Hyperparameters} \\
\cmidrule{2-11}\\
learning\_rate     & 0.18     & 0.18     & 0.36     & 0.18     & 0.15    & 0.27    & 0.15     & 0.39     & 0.37     & 0.27    \\
n\_estimators      & 163      & 10       & 114      & 294      & 320     & 444     & 10       & 280      & 80       & 51      \\
max\_depth         & 6        & 6        & 3        & 6        & 5       & 3       & 4        & 2        & 3        & 5       \\
min\_child\_weight & 10       & 10       & 5.8      & 10       & 10      & 10      & 7.31     & 10       & 10       & 8.65    \\
gamma              & 6.76     & 10       & 4.34     & 9.35     & 10      & 2.97    & 8.23     & 6.84     & 10       & 3.99    \\
subsample          & 0.97     & 0.97     & 1        & 0.97     & 0.51    & 0.95    & 0.53     & 0.61     & 0.65     & 1       \\
colsample\_bytree  & 0.5      & 0.5      & 0.61     & 0.5      & 0.8     & 0.78    & 0.82     & 0.94     & 0.5      & 0.9     \\

\\
 
& \multicolumn{8}{c}{Best Selected Features} \\
\cmidrule{2-11} 
Count   & \textit{19}       & \textit{17}       & \textit{7}        & \textit{19}       & \textit{7}       & \textit{7}       & \textit{13}       & \textit{25}       & \textit{7}        & \textit{7}       \\
\\
Features Indices & \textbf{1}, 5, \textit{6}, 7, \textbf{8}, 15,  21, 22, 24, 25, 26, 36, 40, 49, 53, 55, 56, 58, 60
   &
  \textbf{1}, 3, 5, \textit{6}, 7, \textbf{8}, 13, 21, 23, 28, 29, 31, 39, 48, 53, 55, 56 &
  0, \textbf{1}, \textit{6}, \textbf{8}, 21, 41, 46 &
  \textbf{1}, 2, \textit{6}, 7, \textbf{8}, 15, 21, 22, 24, 25, 28, 31, 32, 39, 40, 51, 56, 59, 60 &
  0, \textbf{1}, 2, 5, \textbf{8}, 14, 56 &
  0, \textbf{1}, \textit{6}, \textbf{8}, 9, 47, 56 &
  0, \textbf{1}, 2, \textit{6}, 7, \textbf{8}, 14, 21, 25, 53, 54, 55, 56 &
  \textbf{1}, 5, \textit{6}, 7, \textbf{8}, 9, 12, 13, 15, 16, 22, 27, 28, 29, 31, 36, 37, 44, 46, 50, 51, 52, 56, 59, 60 &
  0, \textbf{1}, \textit{6}, \textbf{8}, 20, 29, 59 &
  0, \textbf{1}, \textit{6}, 7, \textbf{8}, 24, 60
  \\
  \bottomrule
\end{tabular}
\end{table*}

\subsection{Analysis of the Selected Features} \label{subsec_featureanalysis}
\begin{figure}[!htb]
  \centering
  \includegraphics[width=1.1\linewidth]{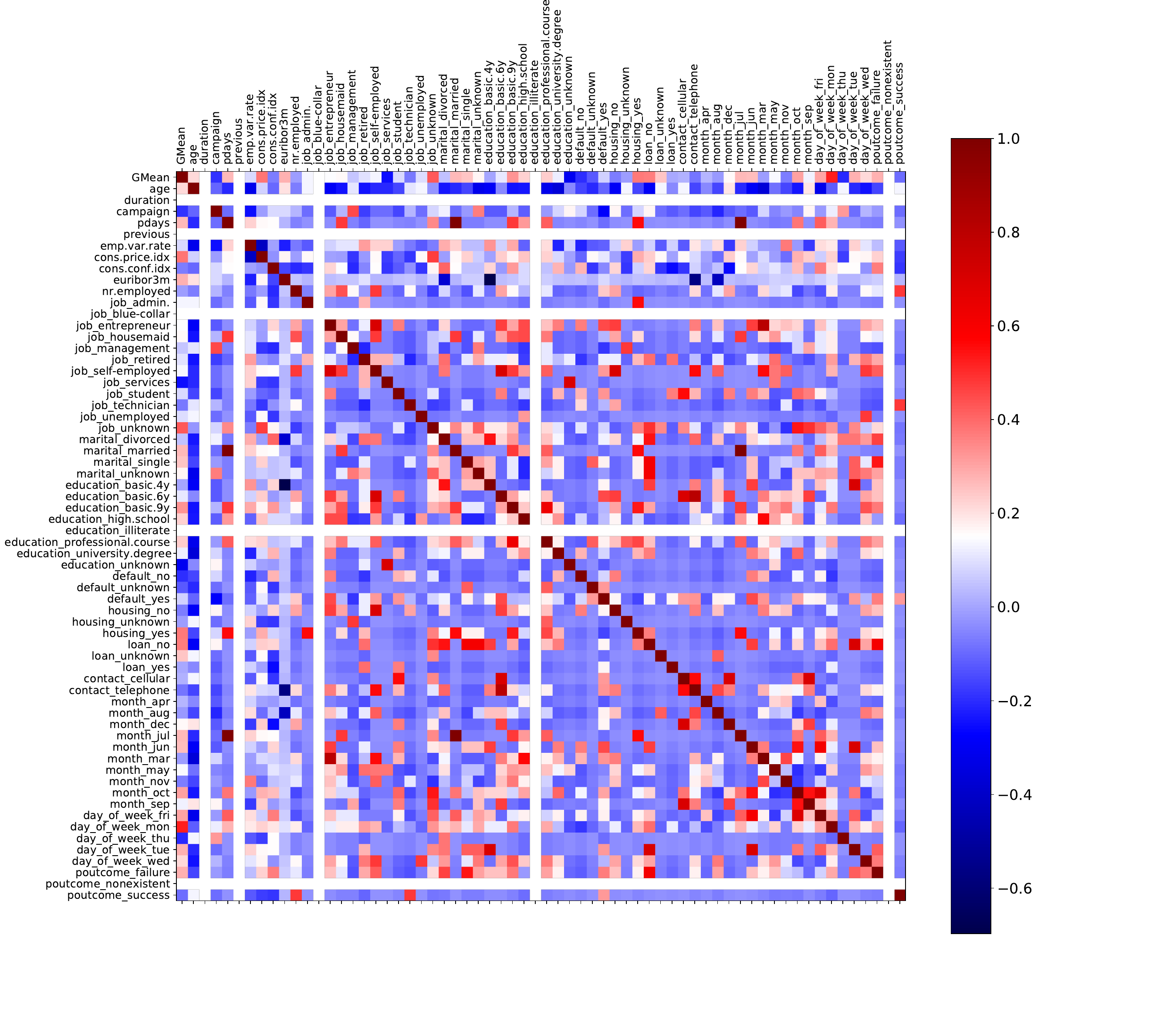}
  \caption{Spearman rank correlation matrix of all the features, in addition to the GMean value}
  \label{fig:CorrAllExpVar}
\end{figure}

The GA achieved a maximum GMean value of 90.21\% with 19 selected features and the best GMean was 89.83\% with seven selected features. In general, the most dominant features are ``1: duration'' and ``8: euribor3m'' followed by ``6: cons.price.idx'' then ``56: day\_of\_week\_mon''. ``duration'' in seconds represents the duration of the last contact with the client,  ``euribor3m'' and ``cons.price.idx'' are socio-economic metrics representing a daily indicator of 3-month ``Euribor'' rate and a monthly indicator of consumer price index respectively, and ``day\_of\_week\_mon'' is the last day of contact with the client. Spearman rank correlation matrix illustrated in Figure \ref{fig:CorrAllExpVar} highlights the correlation of the features based on all the conducted experiments, Table \ref{tab:SpearCorr} lists the GMean correlation coefficients, and Table \ref{tab:featFreq} lists the frequency of the features in all the experiments. The correlation matrix and the frequency of the features give insights into prioritizing the characteristics to be considered in deciding which client to contact in the next marketing campaign.

An interesting observation regarding the significance of the features can be drawn from the correlation matrix. The features are divided into three main groups according to their positive, negative, and neutral influence on the GMean value. The diverging relationships are explained in Table \ref{tab:SpearCorr} in terms of Spearman rank correlation coefficient $r$. The Spearman $r$ values suggest that the Negative and Neutral groups of features are an extra burden in understanding the intentions of the clients. This could be a direction towards studying the significance of telemarketing features and how to re-construct a telemarketing database. 

\begin{table}[!htb]
\centering
\caption{Spearman rank correlation coefficient $r$ of the features in all experiments with the GMean value.}
\label{tab:SpearCorr}
\renewcommand{\arraystretch}{0.6}% Tighter
\begin{tabular}{lc}
\toprule
{Feature} &     GMean \\
\cmidrule(lr){1-1}
\cmidrule(lr){2-2}
\textbf{Positive Correlation}\\
GMean                         &  1.000000 \\
day\_of\_week\_mon               &  0.527395 \\
job\_unknown                   &  0.423456 \\
cons.price.idx                &  0.382481 \\
housing\_yes                   &  0.375691 \\
loan\_no                       &  0.369523 \\
education\_basic.9y            &  0.326174 \\
month\_oct                     &  0.305689 \\
day\_of\_week\_fri               &  0.303669 \\
poutcome\_failure              &  0.281449 \\
euribor3m                     &  0.281059 \\
day\_of\_week\_tue               &  0.281059 \\
marital\_married               &  0.265445 \\
pdays                         &  0.265445 \\
month\_jul                     &  0.265445 \\
month\_jun                     &  0.257546 \\
loan\_unknown                  &  0.249830 \\
marital\_single                &  0.248120 \\
education\_professional.course &  0.237010 \\
education\_high.school         &  0.230655 \\
day\_of\_week\_wed               &  0.212013 \\
age                           &  0.211448 \\
marital\_unknown               &  0.166648 \\
job\_housemaid                 &  0.163087 \\
month\_dec                     &  0.162366 \\
job\_entrepreneur              &  0.151168 \\
month\_may                     &  0.140823 \\
job\_admin.                    &  0.132722 \\
job\_self-employed             &  0.132722 \\
month\_sep                     &  0.128773 \\
education\_university.degree   &  0.120592 \\
education\_basic.6y            &  0.111977 \\
job\_retired                   &  0.102628 \\
job\_unemployed                &  0.093686 \\
job\_student                   &  0.088125 \\
emp.var.rate                  &  0.087515 \\
default\_yes                   &  0.073810 \\
job\_management                &  0.053003 \\
marital\_divorced              &  0.048086 \\
month\_apr                     &  0.027994 \\
contact\_cellular              &  0.023422 \\
loan\_yes                      &  0.011198 \\
education\_basic.4y            &  0.000000 \\
\midrule
\textbf{Negative Correlation}\\
housing\_unknown               & -0.007807 \\
nr.employed                   & -0.012232 \\
month\_mar                     & -0.013914 \\
month\_aug                     & -0.014813 \\
housing\_no                    & -0.061587 \\
cons.conf.idx                 & -0.063511 \\
month\_nov                     & -0.066659 \\
job\_technician                & -0.073389 \\
contact\_telephone             & -0.074211 \\
poutcome\_success              & -0.093686 \\
default\_unknown               & -0.124915 \\
campaign                      & -0.190675 \\
default\_no                    & -0.194803 \\
day\_of\_week\_thu               & -0.202987 \\
job\_services                  & -0.249830 \\
education\_unknown             & -0.302337 \\
\midrule
\textbf{Neutral Correlation}\\
duration                      &       NaN \\
previous                      &       NaN \\
job\_blue-collar               &       NaN \\
education\_illiterate          &       NaN \\
poutcome\_nonexistent          &       NaN \\
\bottomrule
\end{tabular}
\end{table}

\begin{table}[!htb]
\centering
\caption{Selection frequency of features in the best experiments ``A-J'' and in all the experiments.}
\label{tab:featFreq}
\renewcommand{\arraystretch}{0.6}% Tighter
\begin{tabular}{clcc}
\toprule
 \multicolumn{2}{l}{Feature:} & \multicolumn{2}{c}{Experiments} \\
\cmidrule(lr){1-2}
\cmidrule(lr){3-4}

Index & Name                           & A-J &           All         \\
\midrule
1     & duration                       & 10            & 37            \\
8     & euribor3m                      & 10            & 36            \\
6     & cons.price.idx                 & 9             & 19            \\
56    & day\_of\_week\_mon             & 7             & 10            \\
0     & age                            & 6             & 22            \\
7     & cons.conf.idx                  & 6             & 20            \\
21    & job\_unknown                   & 5             & 7             \\
5     & emp.var.rate                   & 4             & 13            \\
60    & poutcome\_failure              & 4             & 5             \\
2     & campaign                       & 3             & 8             \\
15    & job\_retired                   & 3             & 10            \\
22    & marital\_divorced              & 3             & 6             \\
24    & marital\_single                & 3             & 5             \\
25    & marital\_unknown               & 3             & 5             \\
28    & education\_basic.9y            & 3             & 4             \\
29    & education\_high.school         & 3             & 8             \\
31    & education\_professional.course & 3             & 5             \\
53    & month\_oct                     & 3             & 6             \\
55    & day\_of\_week\_fri             & 3             & 5             \\
59    & day\_of\_week\_wed             & 3             & 4             \\
9     & nr.employed                    & 2             & 4             \\
13    & job\_housemaid                 & 2             & 4             \\
14    & job\_management                & 2             & 4             \\
36    & default\_yes                   & 2             & 8             \\
39    & housing\_yes                   & 2             & 3             \\
40    & loan\_no                       & 2             & 2             \\
46    & month\_aug                     & 2             & 5             \\
51    & month\_may                     & 2             & 6             \\
3     & pdays                          & 1             & 1             \\
12    & job\_entrepreneur              & 1             & 2             \\
16    & job\_self-employed             & 1             & 1             \\
20    & job\_unemployed                & 1             & 1             \\
23    & marital\_married               & 1             & 1             \\
26    & education\_basic.4y            & 1             & 2             \\
27    & education\_basic.6y            & 1             & 2             \\
32    & education\_university.degree   & 1             & 3             \\
37    & housing\_no                    & 1             & 2             \\
41    & loan\_unknown                  & 1             & 1             \\
44    & contact\_telephone             & 1             & 3             \\
47    & month\_dec                     & 1             & 2             \\
48    & month\_jul                     & 1             & 1             \\
49    & month\_jun                     & 1             & 2             \\
50    & month\_mar                     & 1             & 3             \\
52    & month\_nov                     & 1             & 5             \\
54    & month\_sep                     & 1             & 2             \\
58    & day\_of\_week\_tue             & 1             & 1             \\
4     & previous                       & 0             & 0             \\
10    & job\_admin.                    & 0             & 1             \\
11    & job\_blue-collar               & 0             & 0             \\
17    & job\_services                  & 0             & 1             \\
18    & job\_student                   & 0             & 3             \\
19    & job\_technician                & 0             & 4             \\
30    & education\_illiterate          & 0             & 0             \\
33    & education\_unknown             & 0             & 2             \\
34    & default\_no                    & 0             & 3             \\
35    & default\_unknown               & 0             & 1             \\
38    & housing\_unknown               & 0             & 1             \\
42    & loan\_yes                      & 0             & 2             \\
43    & contact\_cellular              & 0             & 1             \\
45    & month\_apr                     & 0             & 2             \\
57    & day\_of\_week\_thu             & 0             & 1             \\
61    & poutcome\_nonexistent          & 0             & 0             \\
62    & poutcome\_success              & 0             & 1              \\
\bottomrule
\end{tabular}
\end{table}

\subsection{Ablation Test}
Further analysis of the proposed approach highlights the effect of FS, Parameter Optimization ``PO'', and the selected cost value ``$\lambda$'' on the overall performance (i.e., represented by the GMean).  A set of combinations eliminating one or more of the main parts of the proposed approach (i.e., FS, PO, and cost sensitivity ``C1 or C6'') are confronted with experiment ``A''. ``C1'' means eliminating cost sensitivity (i.e., equal cost of misclassification where ``$\lambda$''=1) and ``C6'' represents considering cost-sensitive classification using the best-selected cost value that was determined by the sensitivity analysis. The baseline for the benchmark judgment is experiment ``A'' having a GMeamn value of 90.21\% (which considers FS, PO, and C6). Table \ref{tbl_ablationtest} describes the GA setup, optimized XGBoost parameters, and the attained performance using each of the configurations. The naming convention of the experiments illustrates the eliminated or included parts of the approach (e.g., ``PO-C6'' means eliminating FS step, utilizing GA to optimize XGBoost parameters, and considering cost sensitivity where ``$\lambda$'' =6). Apparently, none of the combinations was able to outperform the baseline experiment ``A'', and neglecting all the approach components in experiment ``C1'' did not outperform any of the other experiments.

\begin{table*}[!htb]
\centering
\caption{The effect of applying feature selection ``FS'', parameter optimization ``PO'' and the cost value ``$\lambda$'' on the model performance.}
\label{tbl_ablationtest}
\begin{tabular}{l p{5em} p{3em} p{3em} p{5em} p{3em} p{3em} p{3em} p{3em} }
\toprule
Experiment:              & FS-PO-C6 ``A''        & PO-C6        & C6        & FS-PO-C1        & FS-C1       & FS-C6       & PO-C1        & C1       \\
%\cmidrule(lr){1-1}
\cmidrule(r){2-2}
\cmidrule(r){3-3}
\cmidrule(r){4-4}
\cmidrule(r){5-5}
\cmidrule(r){6-6}
\cmidrule(r){7-7}
\cmidrule(r){8-8}
\cmidrule(r){9-9}

& \multicolumn{8}{c}{GA Setup} \\
\cmidrule{2-9}
Perc. to Sel.           & 30\%     & -     & -     & 30\%     & 30\%    & 30\%    & -     & -     \\
Parents               & 100      & 100      & -       & 100      & 100      & 100      & 100       & -       \\
Crossover             & 20\%     & 20\%     & -         & 20\%     & 20\%    & 20\%    & 20\%     & -     \\
Generations          & 30       & 30       & -      & 30       & 30     & 30     & 30       & -       \\
$\lambda$            & 6        & 6        & 6        & 1        & 1       & 6       & 1        & 1        \\
\\
 & \multicolumn{8}{c}{Optimized Hyperparameters} \\
\cmidrule{2-8}\\
learning\_rate	     &	0.18	&	0.15	&	$\star$	&	0.15	&	$\star$	&	$\star$	&	0.19	&	$\star$		\\
n\_estimators	      &	163	&	110	&	$\star$	&	266	&	$\star$	&	$\star$	&	257	&	$\star$		\\
max\_depth	         &	6	&	6	&	$\star$	&	9	&	$\star$	&	$\star$	&	3	&	$\star$		\\
min\_child\_weight	&	10	&	2.74	&	$\star$	&	10	&	$\star$	&	$\star$	&	4.62	&	$\star$		\\
gamma	            &	6.76	&	9.35	&	$\star$	&	9.54	&	$\star$	&	$\star$	&	0.01	&	$\star$		\\
subsample	          &	0.97	&	0.97	&	$\star$	&	0.63	&	$\star$	&	$\star$	&	0.51	&	$\star$		\\
colsample\_bytree	&	0.5	&	0.96	&	$\star$	&	0.5	&	$\star$	&	$\star$	&	0.61	&	$\star$		\\

\\
& \multicolumn{8}{c}{Performance} \\
\cmidrule{2-9} \\
Fitness (GMean)        & 90.21\%  & 89.50\%  & 86.70\%  & 76.76\%  & 76.29\% & 76.29\% & 75.66\%  & 74.60\%  \\

  \\
\bottomrule
\multicolumn{9}{l}{\footnotesize - Not used.} \\
\multicolumn{9}{l}{\footnotesize $\star$ Default XGBoost Parameters which are 0.3, 100, 6, 1, 0, 1, 1 respectively.}

\end{tabular}
\end{table*}

\begin{figure}[!htb]
  \centering
  \includegraphics[width=\linewidth]{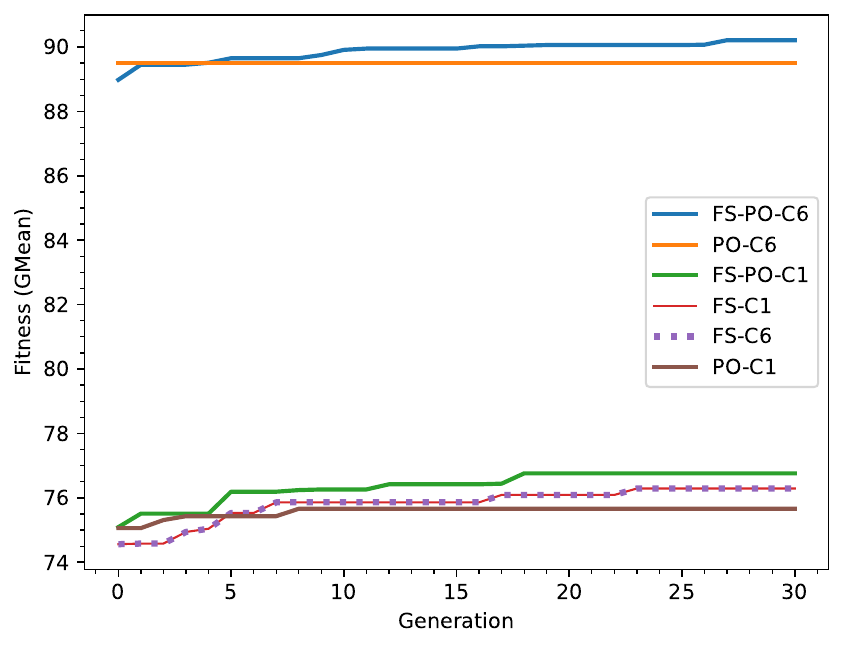}
  \caption{The effect of the ``FS'', ``PO'' and ``$\lambda$'' on attaining the best GMean value using GA.}
  \label{fig:AblationLineChart}
\end{figure}

A closer look at Table \ref{tbl_ablationtest} indicates that the GMean value in experiment ``PO-C6'' is 0.71\%  less than in ``A''. However, the computational complexity of ``PO-C6'' is significantly higher as it considers all the features. Further insights on the effect of the proposed approach steps on the convergence of the GA fitness function are illustrated in Figure \ref{fig:AblationLineChart}. Across the different generations, the bahaviour of the approach indicates that ``FS'', ``PO'', and cost sensitivity together affect positively the performance.

Based on the ablation test, it is essential to consider all the proposed approach components. Noticeably, ``PO'' and cost sensitivity significantly affect the performance. The inclusion of ``FS'' facilitates outperforming and remarkably reduces the complexity of the generated model. ``FS'' and ``PO'' together outperformed having the cost sensitivity disregarded; experiments ``FS-PO-C1'' outperform all non cost-sensitive ones, and ``FS-PO-C6'' outperforms all. 
Understanding the prediction model is better facilitated by less complex models (i.e., smaller optimized trees where the feature subset is relatively small). Human comprehension of less complex models would benefit the marketing decision-making process.

\subsection{Model Validation}
\begin{figure*}[!htb]
  \centering
  \subfigure[Geometric Mean]{\label{fig_CrossValdAJ_GMean}\includegraphics[width=0.45\textwidth]{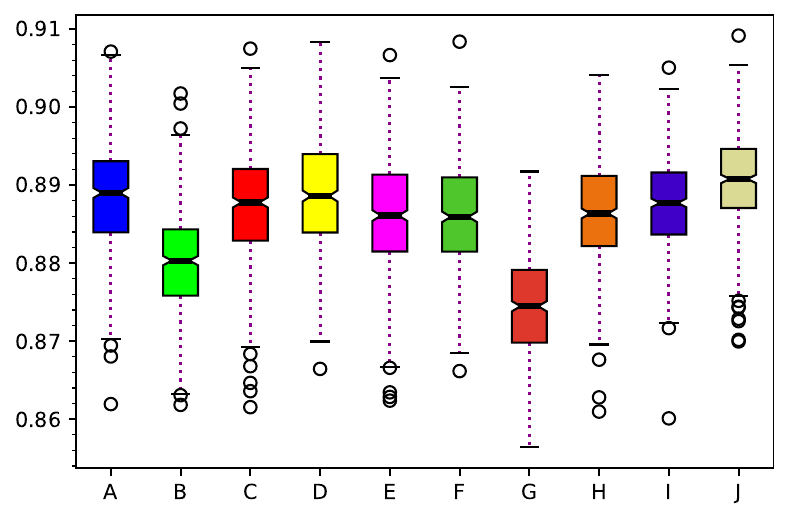}}
  \hspace{1cm}
  \subfigure[Accuracy]{\label{fig_CrossValdAJ_Acc}\includegraphics[width=0.45\textwidth]{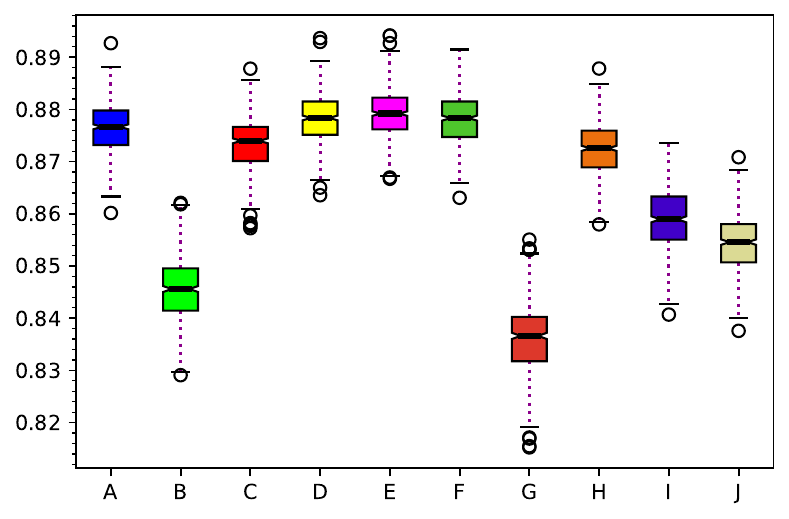}}
  \hspace{1cm}
  \subfigure[TPR]{\label{fig_CrossValdAJ_TPR}\includegraphics[width=0.45\textwidth]{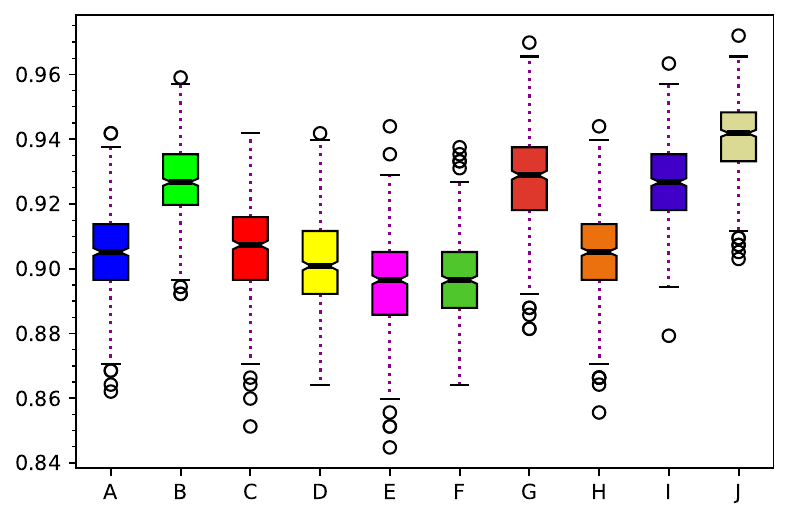}}
  \hspace{1cm}
  \subfigure[FPR]{\label{fig_CrossValdAJ_FPR}\includegraphics[width=0.45\textwidth]{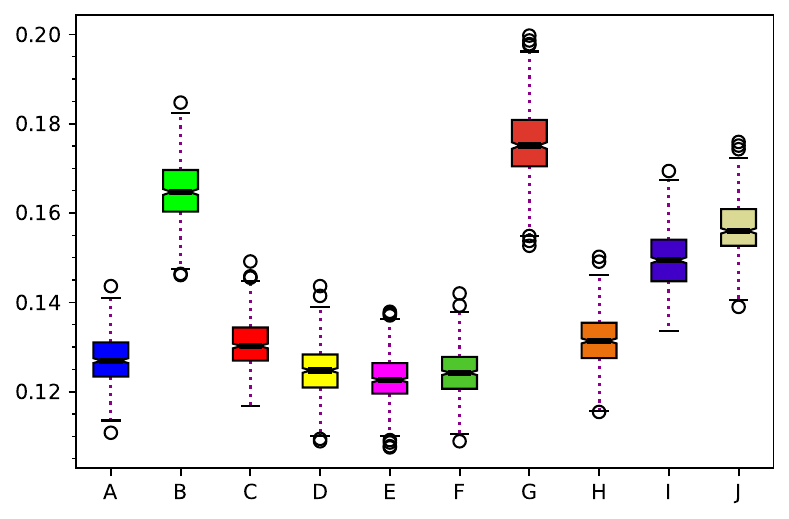}}
  \caption{Performance metrics of ``50$\times$10CV'' using the selected features and optimized parameters in the experiments ``A-J''.}
  \label{fig_CrossValdAJ}
\end{figure*}

A set of 10-fold validation experiments are initialized using the recommended cost values $\lambda$, the optimized XGBoost hyperparameters, and the selected features (i.e., according to the best ten experiments ``A-J'' which are listed in Table \ref{tbl_A-JGA}). Each 10-fold validation experiment based on ``A-J'' is repeated 50 times (50$\times$10CV) to generate 500 models per configuration. Figure \ref{fig_CrossValdAJ} illustrates the performance of the models of each configuration in terms of GMean (Figure \ref{fig_CrossValdAJ_GMean}), Accuracy (Figure \ref{fig_CrossValdAJ_Acc}), TPR (Figure \ref{fig_CrossValdAJ_TPR}), and FPR (Figure \ref{fig_CrossValdAJ_FPR}). The GMean value in the validation experiments is relatively close to the GMean value in the corresponding GA optimization experiments; this indicates the ability of the GA algorithm to find relatively stable models.

The accuracy metric values reassure the concerns in this study and the related works regarding the use of accuracy as the main performance metric in imbalanced data distributions. Based on the performance metrics, the GMean gives better insights regarding the goodness of the prediction model. Usually, in imbalanced data distributions, it is important to boost the accuracy of predicting the minority class without compromising the significance of the majority class. In the case of bank telemarketing, the FPR indicates the ratio of contacts with clients that are less likely to accept an offer. According to Figure \ref{fig_CrossValdAJ_FPR}, the FPR is relatively low in comparison to the TPR; the average of the highest FPR was about 17.5\% in Experiment ``G'' whereas most of the experiments it was less than 13\%. It is highly important to keep the cost value in mind before interpreting the results of the classification algorithm. For example in experiment ``J'' the average TPR was about 94\% but the cost $\lambda$ equals 10; such a scenario calls for a bank marketing domain expert to validate the cost value. If a 10 cost value is justifiable then the prediction model would be suitable for assessing the total cost of marketing efforts in practice. Further details regarding the validation process are presented in Table \ref{tab_detailedCrossValAJ}. The maximum GMean and AUC values were 90.92\% and 95.91\% respectively. Linking the XGBoost parameters (in Table \ref{tbl_A-JGA}) with the results in Figure \ref{fig_CrossValdAJ} and Table \ref{tab_detailedCrossValAJ} illustrates that the higher the ``max\_depth'' the better the performance when $\lambda=6$, except for Experiment ``C''.

\begin{table*}[!htb]
\centering
\caption{Performance and time metrics of the ``50$\times$10CV'' validation experiments.}
\label{tab_detailedCrossValAJ}
%\resizebox{\textwidth}{!}{
\begin{tabular}[t]{clcccccc}
\toprule
  \begin{tabular}[c]{@{}c@{}}Exp\end{tabular} &
  \begin{tabular}[c]{@{}c@{}}GA \\Fitness\end{tabular} &
  \begin{tabular}[c]{@{}c@{}}Selected \\ Features\end{tabular} &
  \begin{tabular}[c]{@{}c@{}}GMean \%\end{tabular} &
  \begin{tabular}[c]{@{}c@{}}AUC \%\end{tabular} &
  \begin{tabular}[c]{@{}c@{}}Training\\ Time\end{tabular} &
  \begin{tabular}[c]{@{}c@{}}Testing\\ Time\end{tabular} \\
  
\midrule 

& \multicolumn{1}{c}{GMean \%} & Count & \multicolumn{1}{c}{Min. : Avg. : Max.} & \multicolumn{1}{c}{Min. : Avg. : Max.} & \multicolumn{1}{c}{Avg. \textit{sec.}} & \multicolumn{1}{c}{Avg. \textit{sec.}}   \\
& & & \multicolumn{1}{c}{(SD)} & \multicolumn{1}{c}{(SD)} & \multicolumn{1}{c}{(SD)} & \multicolumn{1}{c}{(SD)}   \\
% \cmidrule{5} \cmidrule{6}  \cmidrule{7}  \cmidrule{8}
\cmidrule(lr){2-2} \cmidrule(lr){3-3} \cmidrule(lr){4-4} \cmidrule(lr){5-5}  \cmidrule(lr){6-6} \cmidrule(lr){7-7}

A  &
  \textbf{90.21} &
  19 &
  \begin{tabular}[t]{@{}c@{}}86.19 : 88.84 : 90.71\\ (0.007)\end{tabular} &
  \begin{tabular}[t]{@{}c@{}}93.51 : \textbf{94.81} : 95.91\\ (0.003)\end{tabular} &
  \begin{tabular}[t]{@{}c@{}}7.737\\ (0.512)\end{tabular} &
  \begin{tabular}[t]{@{}c@{}}0.107\\ (0.015)\end{tabular} \\
  
B &
  89.97 &
  19 &
  \begin{tabular}[t]{@{}c@{}}86.18 :  88.00 : 90.17\\ (0.006)\end{tabular} &
  \begin{tabular}[t]{@{}c@{}}92.86 : 93.90 : 95.47\\ (0.004)\end{tabular} &
  \begin{tabular}[t]{@{}c@{}}0.573\\ (0.072)\end{tabular} &
  \begin{tabular}[t]{@{}c@{}}0.036\\ (0.008)\end{tabular} \\
    
C &
  89.83 &
  7 &
  \begin{tabular}[t]{@{}c@{}}86.15 : 88.75 : 90.75\\ (0.007)\end{tabular} &
  \begin{tabular}[t]{@{}c@{}}93.38 : 94.54 : 95.46\\ (0.003)\end{tabular} &
  \begin{tabular}[t]{@{}c@{}}2.224\\ (0.298)\end{tabular} &
  \begin{tabular}[t]{@{}c@{}}0.054\\ (0.012)\end{tabular} \\
    
D &
  89.81 &
  19 &
  \begin{tabular}[t]{@{}c@{}}86.64 : 88.87 : 90.83\\ (0.007)\end{tabular} &
  \begin{tabular}[t]{@{}c@{}}93.65 : 94.78 : 95.57\\ (0.003)\end{tabular} &
  \begin{tabular}[t]{@{}c@{}}13.644\\ (0.643)\end{tabular} &
  \begin{tabular}[t]{@{}c@{}}0.136\\ (0.017)\end{tabular} \\
    
E &
  89.75 &
  7 &
  \begin{tabular}[t]{@{}c@{}}86.24 : 88.61 : 90.67\\ (0.007)\end{tabular} &
  \begin{tabular}[t]{@{}c@{}}93.47 : 94.57 : 95.67\\ (0.003)\end{tabular} &
  \begin{tabular}[t]{@{}c@{}}11.8\\ (0.873)\end{tabular} &
  \begin{tabular}[t]{@{}c@{}}0.192\\ (0.031)\end{tabular} \\
    
F &
  89.70 &
  7 &
  \begin{tabular}[t]{@{}c@{}}86.56 : 88.60 : 90.39\\ (0.007)\end{tabular} &
  \begin{tabular}[t]{@{}c@{}}93.46 : 94.52 : 95.42\\ (0.003)\end{tabular} &
  \begin{tabular}[t]{@{}c@{}}9.389\\ (0.493)\end{tabular} &
  \begin{tabular}[t]{@{}c@{}}0.153\\ (0.018)\end{tabular} \\
    
G &
  89.68 &
  13 &
  \begin{tabular}[t]{@{}c@{}}85.64 : 87.43 : 89.17\\ (0.006)\end{tabular} &
  \begin{tabular}[t]{@{}c@{}}92.30 : 93.61 : 94.82\\ (0.004)\end{tabular} &
  \begin{tabular}[t]{@{}c@{}}\textbf{0.504}\\ (0.054)\end{tabular} &
  \begin{tabular}[t]{@{}c@{}}\textbf{0.034}\\ (0.007)\end{tabular} \\
    
H &
  89.65 &
  25 &
  \begin{tabular}[t]{@{}c@{}}86.10 : 88.64 : 90.41\\ (0.007)\end{tabular} &
  \begin{tabular}[t]{@{}c@{}}93.48 : 94.61 : 95.57\\ (0.004)\end{tabular} &
  \begin{tabular}[t]{@{}c@{}}10.875\\ (0.516)\end{tabular} &
  \begin{tabular}[t]{@{}c@{}}0.089\\ (0.013)\end{tabular} \\
    
I &
  89.60 &
  7 &
  \begin{tabular}[t]{@{}c@{}}86.01 : 88.76 : 90.50\\ (0.006)\end{tabular} &
  \begin{tabular}[t]{@{}c@{}}93.46 : 94.42 : 95.27\\ (0.003)\end{tabular} &
  \begin{tabular}[t]{@{}c@{}}1.768\\ (0.201)\end{tabular} &
  \begin{tabular}[t]{@{}c@{}}0.049\\ (0.01)\end{tabular} \\
    
J &
  89.52 &
  7 &
  \begin{tabular}[t]{@{}c@{}}87.00 : \textbf{89.07} : 90.92\\ (0.006)\end{tabular} &
  \begin{tabular}[t]{@{}c@{}}93.61 : 94.71 : 95.58\\ (0.004)\end{tabular} &
  \begin{tabular}[t]{@{}c@{}}1.738\\ (0.183)\end{tabular} &
  \begin{tabular}[t]{@{}c@{}}0.055\\ (0.011)\end{tabular} \\

\midrule
Avg. &
  89.77 &
  - &
  88.56 &
  94.45 &
  6.025 &
  0.089  \\
  
\bottomrule
\end{tabular}
%}
\end{table*}

\begin{figure}[!htb]
  \centering
  \includegraphics[width=1\linewidth]
  {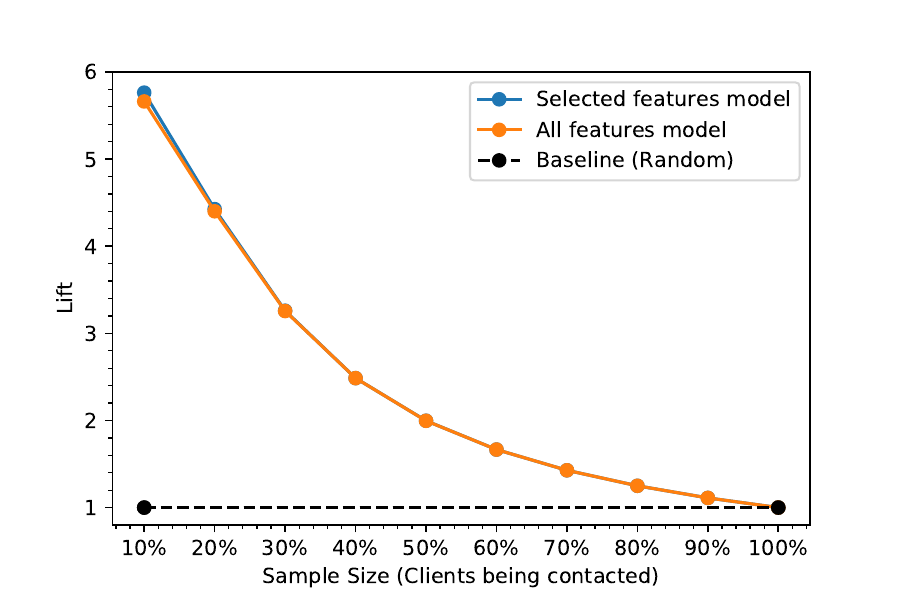}
  \caption{Lift curves of a 10-fold cross-validated XGBoost model, using the selected features (i.e., 30\%) and tuned parameters from experiment ``A''.}
  \label{fig:LiftCurvesCombined}
\end{figure}

\begin{figure}[!htb]
  \centering
  \subfigure[ROC curve using the selected features in exp. ``A''.)]{\label{fig:ROCSelFeat}\includegraphics[width=0.45\textwidth]{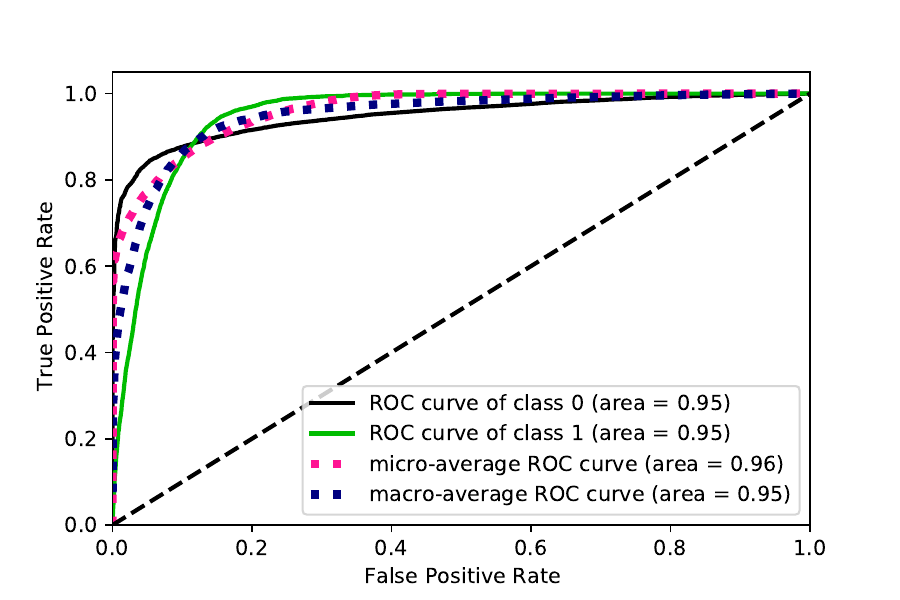}}
  \hspace{1cm}
  \subfigure[ROC curve using all the features.]{\label{fig:ROCallFeat}\includegraphics[width=0.45\textwidth]{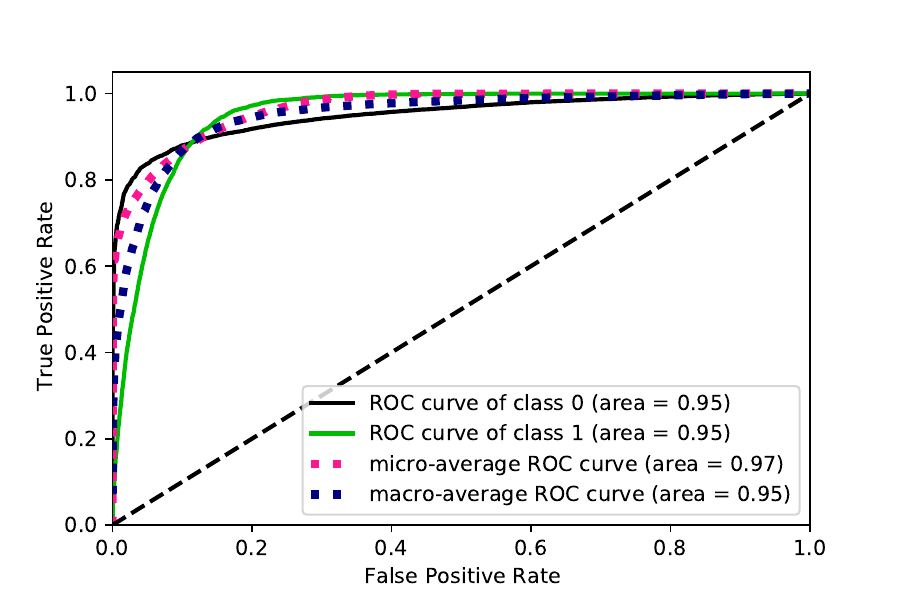}}
  \caption{ROC curves of a 10-fold cross-validated XGBoost model, using the selected features (i.e., 30\%) and tuned parameters from experiment ``A''.}
  \label{fig:ROC}
\end{figure}

Telemarketing teams may create a short list of clients, that are expected to accept a marketing offer, to be contacted. The ability of the prediction model to increase the probability of selecting possibly interested clients is represented by the lift of the classifier, and the diagnostic ability (i.e., the randomness in classification) is illustrated by the ROC; maximum lift and AUC are sought. The selected features (30\% of the 62 encoded features) and optimized hyperparameters in experiment ``A'' (i.e., having the highest GMean and listed in Table \ref{tbl_A-JGA}) were used to generate an XGBoost prediction model. Another model is generated using all the dataset features and the optimized hyperparameters in experiment ``A''. The models were validated using 10-fold stratified cross-validation (10CV). Figure \ref{fig:LiftCurvesCombined} plots the lift curves of both the model with and without feature selection, and Figure \ref{fig:ROC} confronts the ROC curves of the models. 

The ROC curves of the model with the selected features are illustrated in Figure \ref{fig:ROCSelFeat}, and the ROC curves of the second model are plotted in Figure \ref{fig:ROCallFeat}. It is apparent that feature selection reduced the dimensionality of the dataset and maintained competitive lift and AUC values; which supports the objectives of this research in overcoming the challenges imposed by the imbalanced dataset. The lift of the generated model is higher than random selection and the AUC shows that the classifier does not make totally random predictions. 

Based on the lift curve, selecting 10\% of the clients to be contacted using the generated model will make it about 6 times more possible to reach interested clients than by using a random classifier. From another perspective, the AUC indicates a higher probability of the model to rank random interested clients higher than random uninterested more than 95\% of the time. In essence, the lift and ROC curves support the proposed prediction models in terms of leveraging the possibility of reaching the sought clients and lowering the possibility of disturbing uninterested clients. Contacting interested clients and not bothering uninterested ones would be an added value to the bank in terms of its perceived image, and would reduce the telemarketing direct and indirect costs.

The variations in the performance among the different experiments are due to the stochastic nature of the GA search and the randomness of XGBoost model building. Nonetheless, the strengths of both GA and XGboost were combined in this study to select the best subset of features. It is apparent in the results of different experiments that a relatively small subset of the total features could be significantly important in building a competitive prediction model. The outcomes of this study are compared to different related modeling of bank telemarketing attempts in Section \ref{comparisons}.

\subsection{Comparisons} \label{comparisons}
The validated cost-sensitive XGBoost models, tuned by GA, outperform recent related works \cite{ghatasheh2020business,WONG2020112918,9397083,app11199016} in modeling the bank telemarketing process. In essence, the comparison is based on GMean, Type I Error, Type I Error, and Accuracy. Usually, the researchers report part of the performance metrics and used different validation approaches. However, Table \ref{tab:comparision-alg} 
 and Table \ref{tab:comparision2-alg} summarize the performance of several attempts to model part of the bank telemarketing process (i.e., decision-making support in defining profitable clients at the lowest possible financial cost). The total accuracy metric alone is not necessarily an indication of the goodness of the prediction model. Financial cost minimization is manifested by identifying potentially profitable clients who are willing to accept a marketing offer and minimizing the attempts to convince possibly uninterested clients. The trade-off is negotiable as increasing the possibility of identifying a client willing to accept an offer could justify the possible incurred costs of contacting unwilling clients to a certain limit. A domain expert is needed to quantify the exact costs and benefits in money value according to the cost values that are recommended in this study.

% comparisons cost (1)
\begin{table}[!htb]
\centering
\caption{Performance comparison of different classification algorithms.}
\label{tab:comparision-alg}
% \resizebox{\textwidth}{!}{%
% \renewcommand{\arraystretch}{1}% row height
\resizebox{.48\textwidth}{!}{% use resizebox with textwidth
\begin{tabular}{lcccc}
\toprule
\textbf{Algorithm} & \textbf{ Geometric Mean} & \textbf{Type I Error }& \textbf{Type II Error} & \textbf{Accuracy} \\
\midrule\\
\multicolumn{5}{l}{\textbf{Our approach} \textit{(Selected Features)}, Avg. of $50\times10$CV:} \\

A \textit{(f=19)} & 88.84 & 0.096 & 0.127 & 87.65 \\
B \textit{(f=19)}& 88.00 & 0.073 & 0.165 & 84.56 \\
C ~\textit{(f=7)}& 88.75 & 0.094 & 0.131 & 87.35 \\
D \textit{(f=19)}& 88.87 & 0.098 & 0.125 & 87.83 \\
E ~\textit{(f=7)}& 88.61 & 0.105 & 0.123 & 87.92 \\
F ~\textit{(f=7)}& 88.60 & 0.104 & 0.124 & 87.81 \\
G \textit{(f=13)}& 87.43 & 0.073 & 0.175 & 83.62 \\
H \textit{(f=25)}& 88.64 & 0.095 & 0.132 & 87.25 \\
I  ~ \textit{(f=7)}& 88.76 & 0.074 & 0.150 & 85.91 \\
J  ~ \textit{(f=7)}& \textbf{89.07} & \textbf{0.059} & 0.157 & 85.44 \\
\midrule

\multicolumn{5}{l}{\textbf{Other cost-sensitive results from related works:}} \\
Meta-Cost-Multilayer Perceptron \cite{ghatasheh2020business} & 78.93 & 0.192 & 0.229 & 77.48 \\
CostSensitiveClassifier-Multilayer Perceptron \cite{ghatasheh2020business}	&	73.17 &	0.386	& 0.128	& 84.18 \\
Cost-Sensitive Deep Neural Network Ensemble \cite{WONG2020112918}	&	66.2 & 0.295 & 0.378 & n.a. \\
Cost-Sensitive Deep Neural Network \cite{WONG2020112918}	&	57.9 & 0.385 & 0.458 & n.a. \\
AdaCost \cite{WONG2020112918}	&	44.2 & 0.11 & 0.78 & n.a. \\
Meta-Cost \cite{WONG2020112918}	&	55.1 & 0.650 & 0.132 & n.a. \\
\midrule

\multicolumn{5}{l}{\textbf{Benchmark Machine-learning classifiers:}} \\
Multilayer Perceptron \cite{ghatasheh2020business}  & 60.87\% & 0.61 & 0.05 & 88.98 \\
Deep Learning for Multilayer Perceptron \cite{ghatasheh2020business} & 57.86\% & 0.64 & 0.07 & 86.24 \\
J48 \cite{ghatasheh2020business}  & 58.79\% & 0.64 & 0.04 & 88.9 \\
Lib-LINEAR SVM \cite{ghatasheh2020business} & 62.46\% & 0.53 & 0.17 & 78.61 \\
Decision Table \cite{ghatasheh2020business}  & 56.58\% & 0.67 & 0.03 & 89.49 \\
Very Fast Decision Rules \cite{ghatasheh2020business}  & 59.13\% & 0.54 & 0.24 & 72.61 \\
Random Forest Trees \cite{ghatasheh2020business}  & 51.44\% & 0.73 & \textbf{0.02} & 89.82 \\
\midrule

\multicolumn{5}{l}{\textbf{Tuned benchmark machine-learning classifiers:}} \\
XgBoost \cite{9397083} & 78.11 & 0.337 & 0.079 & 90.39  \\
Light GBM \cite{9397083} & 77.83 & 0.352 & 0.066 & \textbf{90.92}  \\
Support Vector Machines \cite{9397083} & 68.30 & 0.487 & 0.091 & 88.46  \\
Gaussian Naïve Bayes \cite{9397083} & 57.49 & 0.643 & 0.074 & 83.93  \\
Random Forest Trees \cite{9397083} & 77.32 & 0.354 & 0.075 & 90.46  \\
\bottomrule
\end{tabular}
}
\end{table}

The GMean of all the experiments in this study is significantly higher than in the related works, which indicates achieving a better balance in modeling the negative and positive classes. For instance in \cite{app11199016} the 12 features, listed in Table \ref{tab:featuresInRelated1},  achieved a maximum total accuracy of 94.39\% but had a low prediction accuracy of the class of interest and relatively high negative class prediction accuracy; 78.94\% in TPR and 96.20\% in TNR respectively. Moreover, non of the models in the related works achieved a competitive prediction accuracy of the positive class; except in \cite{XIE2023108874} but with oversampling. The comparison results did not show any outperforming overall performance, even though some of the related works did not cross-validate the models nor reported a repeated validation approach. The neural network-based models such as in \cite{WONG2020112918} were unable to generate outperforming models nor provide clear feature analysis. Neural network models are usually considered as black-box, which makes them less likely preferred in the decision-aid process; due to the difficulty in interpreting the model and its dynamics. In terms of the comparison with benchmark classifiers, tuning is essential to achieve competitive performance. In \cite{ghatasheh2020business,9397083} the base classifiers failed to model the positive class properly, including XGBoost. It is inferred from the comparison that tuning is essential to initialize the classifiers with proper parameters and not all the features should be used in bank telemarketing model building.

% comparisons others (2)
\begin{table}[!htb]
\centering
\caption{Performance of different related classification algorithms.}
\label{tab:comparision2-alg}
\resizebox{.48\textwidth}{!}{% use resizebox with textwidth
\begin{tabular}{lcccc}
\toprule
\textbf{Algorithm} & \textbf{ Geometric Mean} & \textbf{Type I Error }& \textbf{Type II Error} & \textbf{Accuracy} \\
\midrule\\
% \multicolumn{5}{l}{\textbf{Other machine-learning approaches:}} \\
J48 on Original Dataset \cite{app11199016}	&	75.55	&	0.39	&	0.064	&	90.87	\\
J48 and Feature Selection (f=12) \cite{app11199016}	&	87.14	&	0.211	&	0.038	&	94.39	\\

Naïve Bayes \cite{FENG2022368} & 79.39 & 0.206 & 0.206 & 79.39   \\
Decision Trees \cite{FENG2022368} & 81.83 & 0.169 & 0.190 & 81.82   \\
K Nearest Neighbours \cite{FENG2022368} & 83.43 & 0.185 & 0.146 & 83.48   \\
Logistic Regression \cite{FENG2022368} & 83.17 & 0.175 & 0.161 & 83.18   \\
Support Vector Machine \cite{FENG2022368} & 73.28 & 0.311 & 0.221 & 73.48   \\
Random Forests \cite{FENG2022368} & 85.03 & 0.185 & 0.113 & 85.15   \\
Adaboost \cite{FENG2022368} & 84.52 & 0.166 & 0.143 & 84.55   \\
XGBoost \cite{FENG2022368} & 87.28 & 0.108 & 0.146 & 87.27   \\
Gradient Tree Boosting \cite{FENG2022368} & 87.13 & 0.114 & 0.143 & 87.12   \\
LightGBM \cite{FENG2022368} & 87.13 & 0.120 & 0.137 & 87.12   \\
Artificial Neural Network Ensemble$^\dotplus$ \cite{Article:Moro18} & 81.79 &	0.182 &	0.182 & 81.77   \\
META-DES-AAP$^\star$ \cite{FENG2022368} & 89.39 & 0.074 & 0.137 & 89.39   \\

J48 on Resampled Dataset* \cite{app11199016}	&	86.12	&	0.228	&	0.039	&	94.1	\\
Random Subspace-Multi-Boosting* \cite{XIE2023108874} & \textbf{94.73} & \textbf{0.073} &	\textbf{0.032}  & \textbf{94.7}   \\

\bottomrule
\multicolumn{5}{l}{\footnotesize $\dotplus$ Feature relevance and expert evaluation.} \\
\multicolumn{5}{l}{\footnotesize $\star$ Dynamic Ensemble Selection considering Accuracy and Economic Performance (AP) with Meta-Training} \\
\multicolumn{5}{l}{\footnotesize * Dataset resampling has been applied.}
\end{tabular}
}
\end{table}

The authors in \cite{app11199016} used the ``Wrapped Subset Evaluation'' method to select 12 features in building a ``J48'' trees model. They attained better total accuracy than resampling and using the complete original dataset. Their findings support some of the main results in this study such that a subset of the features set may improve the overall performance of the prediction model. Table \ref{tab:featuresInRelated1} confronts the top 15 features in \cite{XIE2023108874} and the selected 12 Features in \cite{app11199016} with the frequency of the corresponding features chosen in this study. The authors in \cite{XIE2023108874} managed to achieve very competitive performance results but with oversampling. There are many concerns about using oversampling such as the generalization of the generated models. Imposing a synthetic group of instances to overcome the imbalance issue may significantly improve the classifier performance, but it will affect the ability of the model to predict according to a future set of instances. It is usually preferred to avoid altering the original dataset significantly and to validate using a real-world representative sample \cite{ghatasheh2020business}.  

\setlength{\tabcolsep}{1.5pt}
\renewcommand{\arraystretch}{0.9}% Tighter
\begin{table}[!htb]
\centering
\caption{The top 15 features in \cite{XIE2023108874} and the selected 12 Features in \cite{app11199016} confronted with the frequency of the corresponding selected features in the experiments ``A-J''.}
\label{tab:featuresInRelated1}
\begin{tabular}{lll}
\toprule
\multicolumn{3}{l}{Selected Features: }\\
in \cite{XIE2023108874} (\textit{rank}:importance) &  in \cite{app11199016}  &  in ``A-J'' (freq.)  \\
\midrule
age(\textit{15}:0.65) &  age       &   0:age(6)    \\
education(\textit{4}:45.98) &  Education   & 31:education\_professional.course(3) \\
&            & 29:education\_high.school(3) \\
&            & 28:education\_basic.9y(3) \\
&            & 32:education\_university.degree(1) \\
&            & 33:education\_unknown(0)\\
&            & 30:education\_illiterate(0) \\

&            & 27:education\_basic.6y(1) \\
&            & 26:education\_basic.4y(1) \\

contact(\textit{11}:8.31) &  contact     & 44:contact\_telephone(1) \\
&            & 43:contact\_cellular(0) \\

duration(\textit{5}:40.25) &  duration  & 1:duration(10)     \\

pdays(\textit{14}:6.71) &  pdays    & 3:pdays(1)     \\

previous(\textit{9}:12.34) &  previous  & 4:previous(0) \\

poutcome(\textit{10}:10.78) &  poutcome    & 60:poutcome\_failure(4)  \\

NA & emp.var.rate & 5:emp.var.rate(4)   \\

NA & cons.price.idx & 6:cons.price.idx(9)\\

NA & cons.conf.idx  & 7:cons.conf.idx(6) \\

euribor3m(\textit{7}:24.88) &  euribor3m   & 8:euribor3m(10)   \\

NA & nr.employed & 9:nr.employed(2) \\

day\_of\_week(\textit{3}:52.32) &  NA &  56:day\_of\_week\_mon(7) \\

job(\textit{1}:70.77) &  NA & 21:job\_unknown(5) \\
&& 15:job\_retired(3) \\
&& 14:job\_management(2) \\
&& 13:job\_housemaid(2) \\
&& 20:job\_unemployed(1) \\
&& 12:job\_entrepreneur(1) \\
&& 16:job\_self-employed(1) \\
&& 19:job\_technician(0) \\
&& 18:job\_student(0) \\
&& 17:job\_services(0) \\
&& 11:job\_blue-collar(0) \\
&& 10:job\_admin.(0) \\

month (\textit{2}:60) & NA & 53:month\_oct(3) \\
&& 51:month\_may(2) \\
&& 46:month\_aug(2) \\
&& 54:month\_sep(1) \\
&& 52:month\_nov(1) \\
&& 50:month\_mar(1) \\
&& 49:month\_jun(1) \\
&& 48:month\_jul(1) \\
&& 47:month\_dec(1) \\
&& 45:month\_apr(0) \\

campaign(\textit{6}:39.87) &  NA & 2:campaign(3) \\

marital(\textit{8}:17) &   NA & 2:marital\_divorced(3) \\	
&& 24:marital\_single(3) \\
&&  25:marital\_unknown(3)\\	
&& 23:marital\_married(1) \\		

housing(\textit{12}:7.16) &  NA & 39:housing\_yes(2) \\
&& 37:housing\_no(1) \\
&& 38:housing\_unknown(0) \\

loan(\textit{13}:6.97) &   NA & 40:loan\_no(2) \\
&& 41:loan\_unknown(1) \\
&& 42:loan\_yes(0) \\

\bottomrule
\end{tabular}
\end{table}

In summary, the proposed approach in this research managed to outperform the related research in minimizing the misclassification cost of the positive class at the best possible tradeoff. The findings of this research agree with related works that a subset of the dataset features would give insights into the probability of a client accepting a marketing offer. Moreover, one-hot encoding broadens the insights and decomposes the categorical features into corresponding detailed features. The empirical results support the claims regarding the ability of the cost-sensitive model to lower telemarketing costs and describe the dataset features without altering the real-world data. The ultimate goal of this research was to avoid altering the imbalanced real-world data and to seek a potentially realistic prediction model. 

\subsection{Limitations and Implications} \label{implicationslimitations}
This research realized some outperforming and competing models in comparison to the related works on the same dataset. However, some limitations exist which were handled or mitigated. This research considered a single benchmark dataset, which was introduced in \cite{moro2014data}, in all the experiments. The dataset is challenging due to the relatively high imbalance ratio and the existence of some missing values (i.e., missing values provided by the source are denoted as ``unknown''). Stratified and repeated cross-validation of the models, feature encoding, and making the classification algorithm cost sensitive were employed to mitigate the effects of imbalance, minimize the bias in generalization, and avoid any major alterations to the original dataset. The authors in \cite{moro2014data} considered one country and data up to November 2010. In fact, expanding the dataset is not an easy task. This research relies on the results of the related studies on the same dataset in assessing the level of performance improvement. Considering different datasets, building a new dataset, or expanding the dataset would be considered in future works due to time and resource limitations. It is worth mentioning, that socio-economic features are usually specific to a certain community and time which make expanding a dataset even more challenging. The employed classifier (XGBoost) is proven to perform well for the use of imbalanced datasets and outperforms the related research on the same dataset as well. Due to time and resource limitations, this research did not utilize deep learning or highly resource-demanding algorithms but instead, intensive validation and comprehensive reporting of the employed approach results were presented to make it easier to compare with future works.
Despite the limitations and challenges, the implications of this research would benefit the business analytics research area, the bank marketing process, and knowledge-based decision-making in general. Specifically for bank marketing, telemarketers will be equipped with updated knowledge in their communication with the clients which in its turn would improve the customer experience; contacting a wider range of possibly interested clients based on better marketing techniques would stimulate more clients to make a long term deposit, which on its turn will increase the revenues and most probably higher commission for the telemarketers. Less stress on the telemarketers as they would be more confident in their selection of target clients to be contacted and enhances their persuasion abilities. From the telemarketers' point of view, the updated knowledge would make them more engaged and feel satisfied with their achievements; high morale would be another indirect implication. Increased trust between the client and the bank is one of the potential implications that are hard to attain in a short time usually. As a result, positive implications would be on the behaviors of both clients and telemarketers which in its turn will propagate to benefit the organization at the higher levels (e.g., The middle and top management of the bank).
The outcomes of this research create optimistic expectations of more considerations and improvements in future research in the field of bank telemarketing or other domains as well; both in the technical or managerial aspects of the employed approach.

%%%%%%%%%%%%%%%%%%%%%%%%%%%%%%%%%%%%%%%%%%%%%%%%%%%%%%%%%%%
%%%%%%%%%%% Conclusions and Future Directions %%%%%%%%%%%%%
%%%%%%%%%%%%%%%%%%%%%%%%%%%%%%%%%%%%%%%%%%%%%%%%%%%%%%%%%%%
\section{Conclusions and Future Directions} \label{Sec_conclusion}
In light of what is revealed in this study, it is clear that the proposed approach was able to outperform in terms of modeling the telemarketing process. Simultaneous feature selection and hyperparameter optimization using genetic algorithms managed to realize a promising cost-sensitive prediction model. The extensive validation of the prediction model illustrates its ability to mitigate a number of challenges in the analytics of telemarketing data. In essence,  selecting the proper feature subset may significantly improve the performance of the classification model (i.e.,88.84\% and 89.07\%  of GMean using 30\% and 10\% of the features respectively. Furthermore, it is found that it is highly recommended to analyze the effect of the features on the performance as some may negatively correlate with the class labels. From a socio-psychological perspective, revealing aspects of clients would result in a better customer experience, a higher probability to accept an offer, tightened trust, and possibly less job-based stress. However, business domain experts are still needed to quantify the exact financial cost of the marketing campaigns based on the generated models. This research built on the results of previous research work to provide better marketing insights and directs future work toward several related research areas. Such future directions include applying the proposed approach to different telemarketing datasets and applying the same approach to different business domains.

\bibliographystyle{modIEEEtran}
% Loading bibliography database
\bibliography{references}

\begin{IEEEbiography}[{\includegraphics[width=1in,height=1.25in,clip,keepaspectratio]{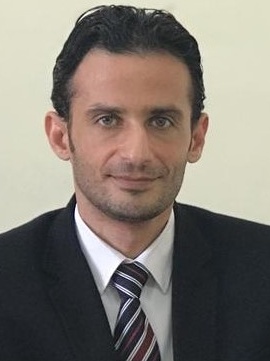}}]{Nazeeh Ghatasheh}  is an Associate Professor at the Information Technology Department of the University of Jordan in Jordan. He holds a B.Sc. degree in computer information systems, which he obtained from The University of Jordan in Amman, Jordan, in 2004. With a strong commitment to academic excellence, he earned merit-based scholarships to pursue his M.Sc. in e-business management and his Ph.D. degree in e-business from the University of Salento in Italy, which he obtained in 2008 and 2011 respectively.

Throughout his academic career, he has held various leadership positions, including Chairperson of the Information Technology and Computer Information Systems Departments, University Liaison Officer for Quality and Development, Dean's Assistant for Quality and Development, Faculty Board Member, and Director of the Computer Center at The University of Jordan in Aqaba, Jordan.

Dr. Ghatasheh has a keen interest in applied scientific research that includes e-Business, Business Analytics, Applied Computational Intelligence, and Data Mining. He is a team member of the funded Erasmus+ KA02 CBHE project named Edu4ALL; which aims at fair inclusion of students with disability in higher education and establishing a technology-based and modern accessibility unit. He is the manager of a funded project by the University of Jordan aimed at realizing an AI-based pronunciation assessment of spoken foreign languages.
\end{IEEEbiography}

\begin{IEEEbiography}[{\includegraphics[width=1in,height=1.25in,clip,keepaspectratio]{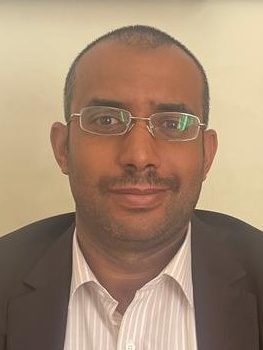}}]{Ismail Altaharwa} received the B.Sc. degree in computer science and its applications from The Hashemite University, Jordan, in 2005, the M.Sc. degree in computer science (emphasis in AI techniques especially computational intelligence and evolutionary computations) from Al-Balqa Applied University, Jordan, in 2008, and the Ph.D. degree in computer science and information engineering (emphasized in machine learning techniques and information security) from the National Taiwan University of Science and Technology, Taiwan, in 2014. 

He is currently an Associate Professor with the Department of Computer Information Systems, The University of Jordan, Aqaba Campus. Altaharwa’s research interests include information security, machine learning, information retrieval, and software engineering. 

Altaharwa is the manager of Edu4ALL project at the University of Jordan. Edu4ALL is an Erasmus+ KA02 CBHE project that aims to encourage the inclusion of students with disability in higher education by modernizing the legalizations and the services provided to them in the HEIs in Palestine and Jordan. And, by establishing an accessibility unit, equipped with the latest tools of assistive technology.
\end{IEEEbiography}

\begin{IEEEbiography}[{\includegraphics[width=1in,height=1.25in,clip,keepaspectratio]{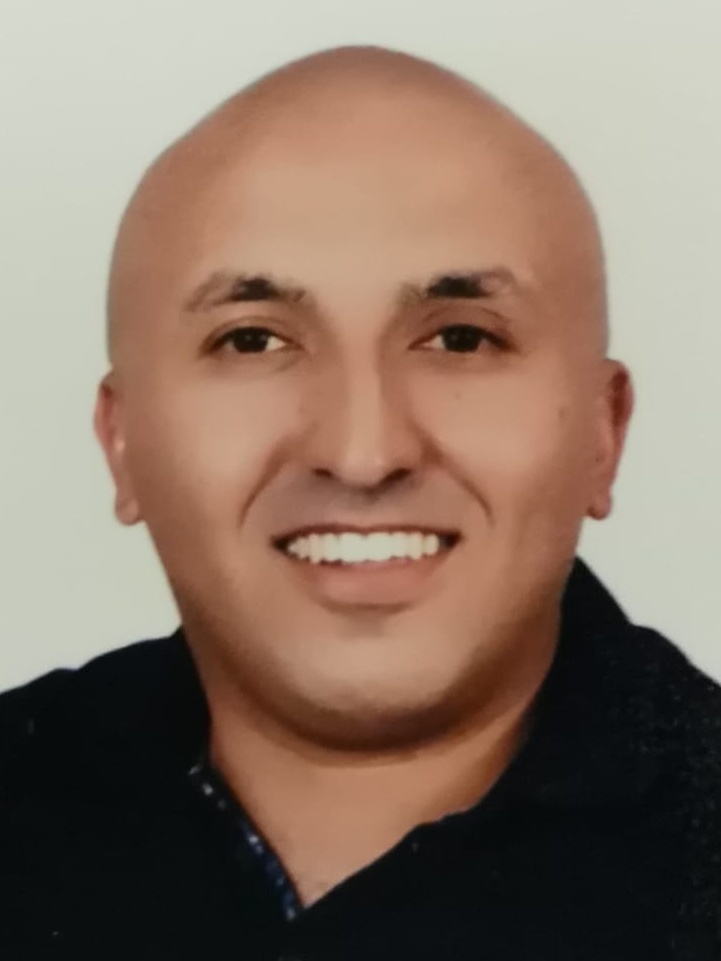}}]{Khaled Aldebei} 
is an Assistant Professor at the Information Technology Department of the University of Jordan in Jordan. He received a B.Sc. degree in Software Engineering from Al-Balqa Applied University, Jordan, in 2006, an M.Sc degree in Computer Science from from Al-Balqa Applied University, Jordan, in 2009, and a Ph.D. degree in Computer Science from the University of Technology, Sydney, Australia, in 2018. 

Throughout his academic career, he has held various leadership positions, including Chairperson of the Information Technology and Computer Information Systems Departments at The University of Jordan in Aqaba, Jordan.

Dr. Aldebei has a keen interest in applied scientific research that includes machine learning, data mining, and natural language processing. He is a team member of the funded Erasmus+ KA02 CBHE project named Edu4ALL; which aims at fair inclusion of students with disability in higher education and establishing a technology-based and modern accessibility unit. He is also a team member of a funded project by the University of Jordan aimed at realizing an AI-based pronunciation assessment of spoken foreign languages.
\end{IEEEbiography}

\EOD

\end{document}